\RequirePackage{fix-cm}
\documentclass[a4paper,11pt]{article}
\usepackage{times}
\usepackage[T1]{fontenc}
\usepackage{upquote}
\usepackage{microtype}
\usepackage{amsmath,amssymb,mathtools}
\usepackage{booktabs}
\usepackage{multirow}
\usepackage{makecell}
\usepackage{longtable}
\usepackage{siunitx}
\usepackage{xcolor}
\usepackage{graphicx}
\usepackage{caption}
\usepackage{subcaption}
\usepackage{xurl}
\usepackage{hyperref}
\usepackage{cleveref}
\usepackage{natbib}
\usepackage{enumitem}
\usepackage{tabularx}
\usepackage{float}
\usepackage{paper-tables}
\usepackage{arxiv-style}

\newcommand{\ptrue}{\textsc{P(True)}}
\newcommand{\pik}{\textsc{P(IK)}}

\newcommand{\datalabel}[1]{\texttt{\small #1}}

\title{On the Limits of Metacognitive\\Monitoring in LLMs}
\author{Dongqi Han \qquad Yifan Yang \qquad Dongsheng Li\textsuperscript{*}\\[4pt]{\fontsize{10.5}{13}\selectfont\color{preprintmuted}Microsoft Research Asia}\\[3pt]{\fontsize{9}{12}\selectfont\color{preprintmuted}\textsuperscript{*}Corresponding author: \href{mailto:Dongsheng.Li@microsoft.com}{Dongsheng.Li@microsoft.com}}}
\date{}
\hypersetup{pdftitle={On the Limits of Metacognitive Monitoring in LLMs},pdfauthor={Dongqi Han, Yifan Yang, Dongsheng Li}}

\begin{document}
\maketitle

\begin{abstract}

Reliable decisions depend on recognizing when an answer may be wrong. In biological cognition, metacognitive monitoring can dissociate from task performance, raising the question of how closely solving and judging are linked in language models. Here we study the confidence reports of four frontier models across 15 benchmarks. High task accuracy can coexist with weak error discrimination: a model solves 97\% of competition mathematics problems while its answer-time confidence ranks correct answers above errors barely better than chance. Confidence separates correct answers from errors more effectively on questions solved by a separate reference model, while review brings limited improvement on reference-hard questions. Aggregate discrimination also rewards ranking correct answers on easy questions above errors on hard ones, which question-only forecasts already do well. Cross-evaluation helps most where the evaluator answered correctly, and errors shared by the two models usually retain high confidence. Hard questions and shared errors remain difficult targets for prompted self-review and peer oversight, even in models with strong problem-solving performance.

\end{abstract}

\section{Introduction}
\label{sec:introduction}

When a person takes an exam, a common strategy is to judge the difficulty of each question and start with those they feel confident they can answer correctly. This strategy relies on metacognition, which has two parts: monitoring and control \citep{nelson1990metamemory}. Metacognitive monitoring is how we judge our own knowledge and performance; metacognitive control uses these judgments to guide what we do next, such as choosing which question to attempt or which answer to check.

Because metacognitive feelings routinely guide how people allocate mental effort \citep{ackerman2017metareasoning}, it is easy to treat such monitoring as a built-in part of intelligence and to assume that any system capable of solving hard problems must also know when it has solved them. On MATH-500, 96.8\% of GPT-5.6 Sol's answers are correct, yet its answer-time confidence ranks a correct answer above a wrong one barely more often than a coin flip would (Section~\ref{sec:solving-is-not-knowing}). Most answers are right, but the confidence reported alongside them gives little guidance about which ones to check. The question is especially pressing in 2026, as large language models (LLMs) increasingly work as agents that write code, run analyses, and act on their own outputs, often without a human checking each step. How well do their confidence judgments identify the answers that need another look?

Previous work shows that LLMs can express useful confidence and judge whether an answer is correct \citep{kadavath2022language,lin2022teaching,tian2023just}. Yet they often fail to locate errors in reasoning traces \citep{tyen2024errors}, and intrinsic self-correction remains unreliable without external feedback \citep{huang2024cannot,kamoi2024survey}. Confidence may carry information about a question's difficulty as well as the quality of a particular answer. A model may catch a slip on a familiar problem while remaining confident in a plausible answer to an unfamiliar one. This makes the distribution of errors central: which mistakes does confidence reveal, and which become visible only on review?

We put these questions to four frontier models on 15 benchmarks spanning knowledge, mathematics, reasoning, code, multimodal, long-context, and generalist tasks, with 38{,}238 items per model. We compare confidence reported from the question alone, alongside the answer, and during a fresh-context review of the completed answer and its supporting solution. Each later judgment addresses the same candidate answer, allowing us to isolate changes in evaluation. We then cross solvers with evaluators to ask what another model adds to self-review.

Our findings reveal substantial variation in error discrimination even at similar levels of task accuracy. Reviewing fixed answers can improve or weaken that discrimination. The clearest shared pattern concerns difficulty: across all four models, confidence distinguishes correct answers from errors less effectively on questions missed by a separate reference model. These questions account for a substantial share of errors in the three-setting comparison. Aggregate scores also reward separating easy questions from hard ones, a distinction that question-only forecasts already capture. Review offers limited improvement within the hard questions themselves.

Peer evaluation reveals a related dependence on the evaluator's own answer. When GPT-5.6 Sol and Gemini 3.8 Flash evaluate each other's answers, they give higher confidence to a wrong option that matches their own than to a correct option they failed to choose. The same evaluator can assign high confidence to two mutually exclusive answers when judging them separately. Cross-evaluation is more useful where the evaluator solved the question correctly; where both models made the same mistake, the error usually survives both judgments. Agreement can make an answer reassuring without making it right.

Holding answers fixed, comparing judgments within difficulty groups, and crossing solvers with evaluators reveal what confidence contributes to decisions about checking and deferral. We focus on the information these judgments provide for deciding when to revisit an answer or seek another evaluator. Acting on an error requires a signal that brings it to attention and a way to obtain a better answer. The usefulness of self-review and peer oversight depends on both.

\section{Results}
\label{sec:results}

\subsection{Evaluation Protocol}
\label{sec:evaluation}

We evaluated four frontier language models (GPT-5.6 Sol, Qwen3.8-27B, Gemini 3.8 Flash, and Grok 4.6) on the same 38{,}238 items from 15 public benchmarks in seven domains, spanning knowledge, mathematics, reasoning, code, multimodal, long-context, and generalist tasks (Appendix~\ref{app:protocol}). Twelve benchmarks use deterministic grading; Humanity's Last Exam and LiveMathBench use a GPT-5 mini judge, and SimpleQA uses GPT-4.1 supplemented by exact numerical comparisons. Each model reported its confidence on an integer scale from 0 to 100 in three settings, which correspond to the prospective, concurrent, and retrospective judgments of metacognition research, made before, together with, and after a response \citep{fleming2026integrative}. All three reports are scored against the correctness of the same recorded answer.

\textbf{\emph{Pre-answer confidence}} is a question-only operationalization of prospective monitoring \citep{fleming2026integrative}. The model receives the task, including any options and images, but no candidate answer or solution, and estimates its probability of answering correctly; it may reason internally but is asked not to reveal a proposed answer or solution. The name denotes this information condition, not the collection order: forecasts were collected after the answers had been recorded, without exposing them. \textbf{\emph{Answer-time confidence}} is reported in the same generation as the answer and its written supporting solution. Its closest counterpart is peri-decision wagering, in which a response and a confidence judgment are elicited together \citep{fleming2026integrative}; it can depend on the reasoning and response tokens that precede it, but not on text generated later. \textbf{\emph{Post-answer confidence}} is a retrospective review \citep{nelson1990metamemory}. In a fresh context, the model receives the question with its own completed answer and written supporting solution, but not its original confidence or hidden solve trace; instructed not to revise the answer, it estimates the probability that this fixed answer will be graded fully correct. The prompt identifies the answer as the model's own, so the fresh context does not make its source anonymous. Pre-answer and Post-answer confidence correspond to the question-level \pik\ and answer-level \ptrue\ of \citet{kadavath2022language}.

The three settings are distinct elicitation conditions, not successive snapshots of one reasoning trace. For each model--benchmark pair, we compare them on the items with valid reports in all three: 99.5\% of items for GPT-5.6 Sol, 93.3\% for Qwen3.8-27B, 77.7\% for Grok 4.6, and 66.7\% for Gemini 3.8 Flash. Most exclusions are responses that violated the required JSON-only output format; they count as errors in task accuracy but do not yield valid confidence reports under the strict output contract (Appendix~\ref{app:self-evaluation-results}). Appendix~\ref{app:confidence-prompts} gives the prompts and information available to each report; Appendix~\ref{app:retention-recovery} reports coverage by benchmark and difficulty tier and deterministic format-recovery results.

\subsection{Metacognitive monitoring atlas}
\label{sec:metacognitive-monitoring-atlas}

\begin{figure}[p]
  \centering
  \includegraphics[width=5.5in]{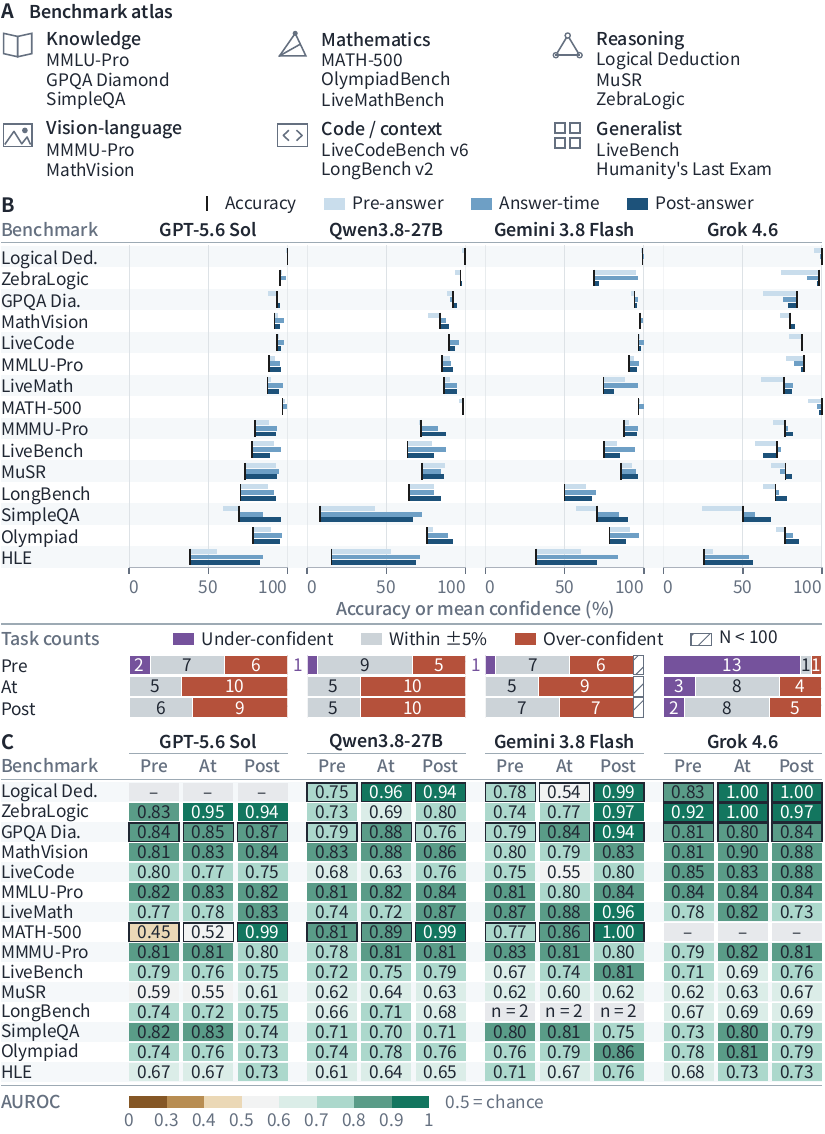}
  \caption{\textbf{Metacognitive monitoring across 15 benchmarks.} \textbf{A}~Benchmarks grouped by task. \textbf{B}~Black ticks mark accuracy; blue bars extend to mean confidence (right: overconfidence; left: underconfidence; darkest: Post-answer). Strips count tasks below, within, or above a $\pm5\%$ band around accuracy (five percentage points); hatching marks tasks with $N<100$ analyzed answers, left unclassified. Counts describe mean bias, not item-level calibration. \textbf{C}~Tile colors encode within-task AUROC, not mean confidence ($0.5$: chance). Outlines mark fewer than 20 correct or 20 incorrect answers (unstable estimates). --:~all answers correct, so AUROC is undefined. $n=2$:~too few to estimate AUROC. B and C use identical answers with valid reports at all three confidence stages. Most missing reports violate the JSON-only output format; these count as errors in benchmark accuracy but are excluded here (Appendix~\ref{app:self-evaluation-results}).}
  \label{fig:confidence-stages}
\end{figure}

Figure~\ref{fig:confidence-stages}B compares each model's accuracy with its mean confidence. At answer time, mean confidence exceeded accuracy by more than five percentage points on most tasks for three of the four models. An average, however, cannot show whether confidence is well placed on individual answers. A model that reports 95\% confidence on every question and answers 95\% correctly matches its accuracy on average, yet gives every mistake the same high score.

What matters is whether mistakes receive lower confidence than correct answers. A confidence threshold turns this into a decision: answers at or above it are endorsed, and answers below it are rejected. With mistakes as the positive class, a rejected mistake is a true positive (TP), an endorsed mistake a false negative (FN), a rejected correct answer a false positive (FP), and an endorsed correct answer a true negative (TN). Sweeping the threshold traces the receiver operating characteristic: the fraction of mistakes caught, $\mathrm{TP}/(\mathrm{TP}+\mathrm{FN})$, against the fraction of correct answers discarded, $\mathrm{FP}/(\mathrm{FP}+\mathrm{TN})$. The area under this curve (AUROC) equals the fraction of correct--incorrect pairs from the same model and benchmark in which the correct answer receives the higher confidence, with ties counting half; $0.5$ means chance-level ranking, as in the constant-confidence example, and $1$ means that every correct answer ranks above every mistake. In the terms of metacognition research, the gap between mean confidence and accuracy measures metacognitive bias, whereas AUROC measures metacognitive sensitivity \citep{fleming2014measure}. Figure~\ref{fig:confidence-stages}C maps within-task AUROC in all three settings on the same answers as panel~B; within each row the settings share one set of items, although the sets can differ across models. Does a model that solves a task well also know which of its answers are wrong?

\clearpage
\subsection{Solving is not knowing}
\label{sec:solving-is-not-knowing}

\begin{figure}[htb]
  \centering
  \includegraphics[width=5.5in]{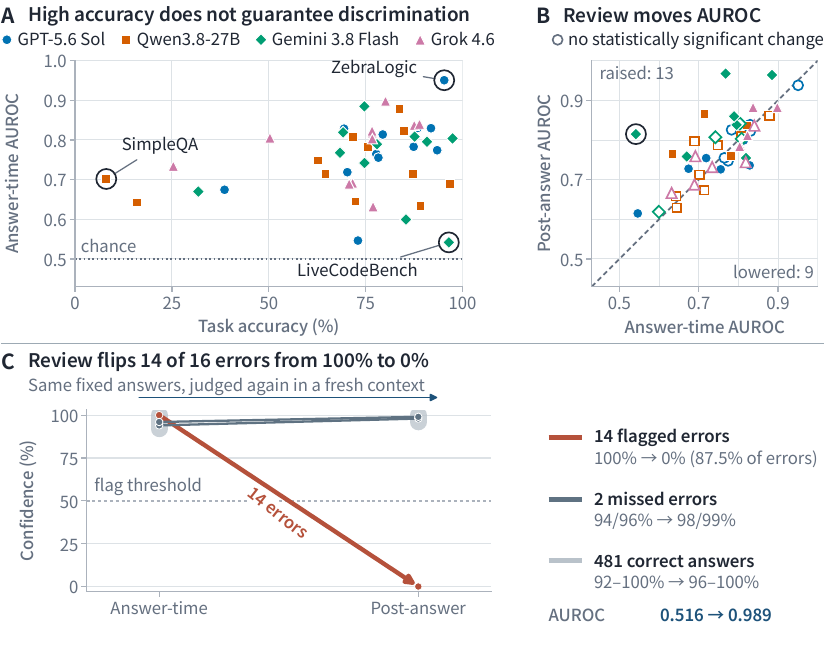}
  \caption{\textbf{Solving is not knowing.} \textbf{A}~Task accuracy and answer-time AUROC for the 46 model--benchmark pairs with at least 100 answers, including at least 20 correct and 20 incorrect ones; each pair uses the answers with valid answer-time and post-answer confidence. Rings mark the pairs discussed in the text; the dotted line marks chance ranking. \textbf{B}~Answer-time and post-answer AUROC on the same answers. Hollow markers: the 95\% bootstrap interval of the difference includes zero (2{,}000 resamples of correct and incorrect answers); corner counts give the pairs whose interval lies entirely above (raised) or below (lowered) zero. The dashed line marks equal AUROC. \textbf{C}~The same 497 fixed GPT-5.6 Sol answers on MATH-500, reviewed in a fresh context; 16 errors fall below the eligibility threshold for panels A and B. One trajectory groups 14 errors falling from 100\% to 0\%, including 13 with visible answer--solution disagreements (Appendix~\ref{app:math500-content}). The other two receive 98--99\% on review. Vertical bars show ranges for the 481 correct answers. The dotted line marks the 50\% flagging threshold.}
  \label{fig:solving-vs-knowing}
\end{figure}

In \emph{direct-access} accounts of metacognition, confidence draws on the evidence used to produce an answer \citep{fleming2026integrative}. Task performance and monitoring can nevertheless dissociate: in rodents and humans alike, prefrontal lesions can impair metacognition while leaving task performance intact \citep{fleming2026integrative}. We find an analogous dissociation in frontier language models: high solving accuracy does not guarantee that the confidence reported alongside an answer distinguishes successes from errors. Of GPT-5.6 Sol's 497 MATH-500 answers with both answer-time and post-answer confidence, 481 were correct (96.8\%), yet its answer-time confidence ranked correct answers above errors barely more often than chance would (AUROC $=0.516$ across its 16 errors; Figure~\ref{fig:solving-vs-knowing}C). It gave the maximum confidence of 100 to 14 of its 16 errors, nearly the same share as among its correct answers (430 of 481).

High accuracy alone does not explain this weak discrimination. Estimates of metacognitive sensitivity are easily confounded with task performance, which is why metacognition research often expresses sensitivity relative to performance, as the metacognitive efficiency meta-$d'/d'$ \citep{fleming2014measure,fleming2026integrative}. Open-ended answers define no $d'$, so we instead compare tasks that the model solves about as well. On ZebraLogic, the only other benchmark GPT-5.6 Sol solved about as often (95.3\%), its AUROC reached 0.951. The dissociation is not confined to one model or benchmark. Among the 46 model--benchmark pairs with enough correct and incorrect answers to estimate AUROC, accuracy and answer-time AUROC were weakly correlated overall (Spearman $\rho=0.28$), with substantial variation across individual models ($\rho$ from $-0.05$ to $0.67$). The 18 pairs with at least 80\% accuracy spanned the full range observed across all pairs, from 0.54 to 0.95: Gemini 3.8 Flash solved 96.5\% of LiveCodeBench v6 problems yet ranked its answers barely above chance (0.542). All five pairs below 50\% accuracy had AUROCs of 0.64--0.73, among them Qwen3.8-27B on SimpleQA, which answered only 8.1\% of questions correctly yet reached 0.702 (Figure~\ref{fig:solving-vs-knowing}A). Similar success rates can conceal very different monitoring abilities.

Our design permits a sharper test, one that holds the answers fixed. Post-answer confidence re-evaluates the very answers that answer-time confidence accompanied, so changes in AUROC reflect new judgments rather than revised answers. Review moved AUROC beyond sampling error in 22 of the 46 pairs, raising it in 13 and lowering it in 9, by as much as $+0.27$ and $-0.09$ (Figure~\ref{fig:solving-vs-knowing}B); the largest gain lifted Gemini 3.8 Flash on LiveCodeBench v6 from 0.542 to 0.815. On MATH-500, review lifted GPT-5.6 Sol from near chance to 0.989, and how it did so is telling (Figure~\ref{fig:solving-vs-knowing}C). Fourteen of its 16 errors fell from 100\% confidence straight to 0\%, while the remaining two and every correct answer were rated 96--100\%: review gave either a categorical rejection or a near-certain endorsement. In 13 of these 14 cases, the written solution already contradicted the answer field, and the review pointed to that discrepancy (Appendix~\ref{app:math500-content}). Here, review largely caught inconsistencies already visible in the completed response. These results reveal a dissociation between solving and judging in language models. Yet high overall discrimination need not mean that confidence detects errors within hard questions: it can also reward separating easy questions from hard ones.

\subsection{Difficulty is not error}
\label{sec:difficulty}

\begin{figure}[htb]
  \centering
  \includegraphics[width=5.5in]{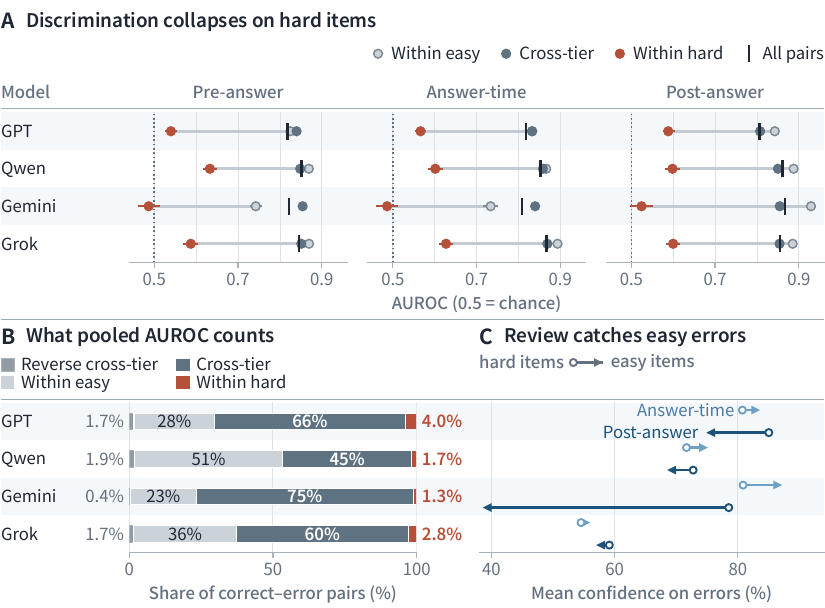}
  \caption{\textbf{Error discrimination remains weak on hard questions.} An item is \emph{easy} if the reference model, Gemini 3.7 Flash, answered it correctly and \emph{hard} otherwise. Rows show GPT-5.6 Sol, Qwen3.8-27B, Gemini 3.8 Flash and Grok 4.6, in that order. All panels use the answers of Figure~\ref{fig:confidence-stages} that have valid reports in all three settings and a reference label (25{,}475--37{,}824 per model), pooled across the 15 benchmarks. \textbf{A}~AUROC in each confidence setting for three pair types: both answers on easy items (within easy), both on hard items (within hard), and a correct answer on an easy item versus an error on a hard item (cross-tier). Grey lines join the within-easy and within-hard values, and ticks mark AUROC over all correct--error pairs. Whiskers are 95\% percentile intervals from 2{,}000 item resamples within benchmark and tier. Pooled-AUROC intervals lie within $\pm$0.01 of their estimates and are omitted; dotted lines mark chance ranking (0.5). \textbf{B}~Share of each pair type among all correct--error pairs; the fourth type, reverse cross-tier, pairs a correct answer on a hard item with an error on an easy item (0.4--1.9\%). Pooled AUROC is the share-weighted mean of the four pair-type AUROCs. \textbf{C}~Mean confidence on incorrect answers, from hard items (open circle) to easy items (arrowhead).}
  \label{fig:difficulty-and-reevaluation}
\end{figure}

Which mistakes can a model recognize, and which stay hidden? Difficulty is the obvious suspect: a model might notice slips on familiar questions yet miss errors on questions it finds difficult. Testing this requires a measure of difficulty that does not depend on the target model's own correctness; otherwise the hard tier would contain only errors. We therefore let an outside model decide: a question is \emph{easy} if Gemini 3.7 Flash answered it correctly and \emph{hard} otherwise. With each model's answers from Figure~\ref{fig:confidence-stages} pooled across the 15 benchmarks, all four models tell the same story (Figure~\ref{fig:difficulty-and-reevaluation}A). Within easy questions, answer-time confidence ranks correct answers above errors with an AUROC of 0.73--0.89. Within hard questions, the AUROC falls to 0.49--0.63; for Gemini 3.8 Flash (0.486), it is close to chance, and no confidence setting lifts it above 0.64 for any model. This matters because of where the errors concentrate: hard questions account for only 14--18\% of each model's analyzed answers but 47--77\% of its errors (ranges across the four models; Table~\ref{tab:difficulty-error-concentration}). On these questions the models are right 10--26\% of the time, yet their mean confidence stays at 57--83\%. Human judges are too sure in the same place, on the questions they answer worst \citep{lichtenstein1977know}. The difficulty gradient persists when questions are instead grouped by how many of two other models, both from outside the target's family, solved them, and hard-question AUROC falls below easy-question AUROC in 33 of 36 comparisons within individual benchmarks (Appendix~\ref{app:difficulty-robustness}). Repeating the matched comparison after leaving out each benchmark in turn preserves the easy--hard ordering (Appendix~\ref{app:label-sensitivity}).

Why, then, do these models achieve pooled AUROCs of 0.81--0.87? Pooled over benchmarks, AUROC is the fraction of all correct--incorrect pairs in which the correct answer receives the higher confidence. Pairs drawn from two hard questions are rare, 1.3--4.0\% of the total, because so few hard questions are answered correctly (Figure~\ref{fig:difficulty-and-reevaluation}B); with the other pair-type AUROCs held fixed, perfect ranking within them would raise pooled AUROC by only 0.006--0.017. Far more pairs pit a correct answer on an easy question against an error on a hard one: these \emph{cross-tier} pairs make up 45--75\% of all pairs, the largest share for three of the four models. They can be won without judging either answer, merely by being less confident on harder questions, and the models win 83--87\% of them. For Gemini 3.8 Flash, pooled AUROC (0.809) even exceeds the AUROC within either tier (0.734 and 0.486), and the per-task AUROCs of Figure~\ref{fig:confidence-stages}C likewise credit a sense of difficulty wherever a task mixes easy and hard questions. A question-only forecast achieves similar discrimination. Pre-answer confidence, elicited without any answer in view, wins cross-tier pairs about as often as answer-time confidence does (0.84--0.85 versus 0.83--0.87), and its pooled AUROC comes within 0.03 of the answer-time value for every model; within either tier, confidence reported with the answer differs from this question-only forecast by at most 0.041 in AUROC.

These similar averages do not make the reports interchangeable. In held-out prediction, adding either Answer-time or Post-answer confidence to Pre-answer confidence improves overall discrimination for every model. Post-answer confidence contributes more additional ranking information on easy questions than on hard ones (Appendix~\ref{app:incremental-confidence}).

Answer-time confidence rose with ease for both correct answers and errors. Errors received $+1.5$ to $+6.3$ percentage points more confidence on easy questions than on hard ones (Figure~\ref{fig:difficulty-and-reevaluation}C; Appendix~\ref{app:difficulty-robustness}). In humans, stronger evidence can raise confidence in correct answers and lower it in errors, a folded X; when the judge can instead tell how hard the task is, confidence rises for both \citep{fleming2026integrative}. Answer-time confidence follows this latter pattern. Review changed the relationship between ease and error confidence. Easy-question errors then received less post-answer confidence than hard-question errors: 4.2--10.2 points less for GPT-5.6 Sol and Qwen3.8-27B and 40.0 [37.1, 42.7] points less for Gemini 3.8 Flash (brackets give 95\% bootstrap intervals). Grok 4.6 showed a smaller between-tier difference ($-1.4$ [$-2.92$, $-0.01$] points). Gemini 3.8 Flash, whose review raised pooled AUROC the most (0.809 to 0.867), cut mean confidence in easy-question errors from 87.2\% to 38.6\% while barely touching hard-question errors (80.9\% to 78.5\%). Within easy questions its AUROC rose from 0.734 to 0.929; within hard questions, only from 0.486 to 0.524 ($+0.038$ [0.012, 0.065]). For the other three models, review moved within-hard AUROC by $-0.027$ to $+0.022$ and increased mean confidence in hard-question errors. A second look can catch slips on questions the model can solve. Hard-question errors remain highly rated, echoing the coupling of weak skill and weak self-assessment described in humans \citep{kruger1999unskilled}.

\subsection{Agreement is not correctness}
\label{sec:crossed}

If a model cannot recognize its own mistakes, can another model? Evaluators can favor their own generations \citep{panickssery2024recognize}, while a plausible wrong answer may also persuade another evaluator. We therefore had GPT-5.6 Sol and Gemini 3.8 Flash each rate both models' fixed answers, every time in a fresh context and under a source-neutral prompt, reporting post-answer confidence (Appendix~\ref{app:crossed-confidence-prompts}). The primary analysis covers 18{,}112 items from nine objectively graded tasks, on which the two models are similarly accurate (89.1\% and 88.1\%; Appendix~\ref{app:crossed-population}). Both models erred on 1{,}318 items. Averaging across tasks, the self-source advantage $h$---how much more an evaluator trusts its own wrong answers than the other model's---was $+3.3$ [0.6, 6.0] percentage points (brackets give 95\% bootstrap intervals). On matching wrong answers, the item-averaged advantage was only $+1.1$ [0.5, 1.7]; when the answers differed, GPT favored its own by $+16.5$ [13.3, 19.8] points, whereas Gemini's preference depended on the task (Appendix~\ref{app:crossed-joint}). The larger rating differences arise when the models disagree on the answer.

\begin{figure}[htb]
  \centering
  \includegraphics[width=5.5in]{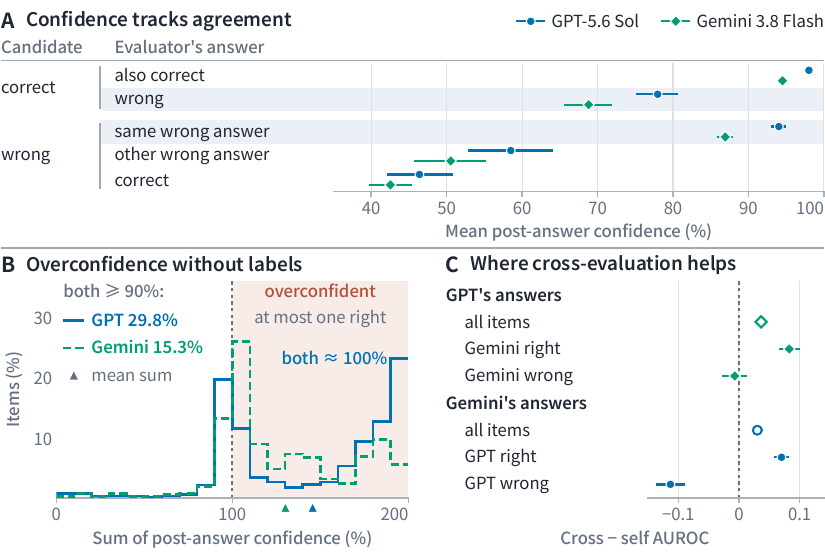}
  \caption{\textbf{Crossed evaluation tracks agreement with the evaluator's own answer.} Models rate fixed answers in fresh contexts under the same source-neutral prompt (author identity withheld). Post-answer confidence is the reported probability that a candidate is fully correct; colors identify the evaluator. \textbf{A}~Mean confidence in the other model's answer by candidate and evaluator correctness (12{,}916 multiple-choice items). Shading contrasts missed correct answers with matching errors. \textbf{B}~Sum of one evaluator's confidence in the models' different options (1{,}017 items; right-closed 10-point bins). Sums above 100\% (shaded) imply overconfidence: at most one option is correct. The legend reports the share with both ratings at least 90\%; triangles mark mean sums. \textbf{C}~Cross minus self AUROC on fixed candidates (nine tasks, including the four multiple-choice tasks in A and B; 18{,}112 items): self uses the answer-source model as evaluator; cross uses the other model. Open markers pool all items; filled markers split by cross-evaluator correctness. Positive differences favor cross-evaluation. Error bars in A and C are 95\% paired bootstrap intervals (5{,}000 item resamples; task list and methods in Appendix~\ref{app:crossed-population}).}
  \label{fig:ce}
\end{figure}

Ratings track agreement with the evaluator's earlier answer. On multiple-choice tasks, a wrong answer identical to the evaluator's own received 94.0\% from GPT and 86.9\% from Gemini, more than a correct answer the evaluator had itself missed (78.0\% and 68.8\%; differences $+16.1$ [13.2, 19.1] and $+18.1$ [14.8, 21.6] points; Figure~\ref{fig:ce}A). This contrast persists on items an external model answered correctly, and rating differences remain after stratifying by external-model correctness (Appendix~\ref{app:crossed-robust}). The same evaluator can also give high confidence to two mutually exclusive answers when each is presented separately. On the 1{,}017 multiple-choice items where the models chose different options, GPT rated both options at least 90\% likely to be correct on 29.8\% [27.2, 32.5] of items, and Gemini on 15.3\% [13.2, 17.4] (Figure~\ref{fig:ce}B). At most one candidate in each pair can be correct. GPT's mean sum of 1.46 therefore implies that its probabilities overstate accuracy by at least 23 percentage points for GPT and 15 for Gemini, without a single label. Read on their own, conflicting solutions can both persuade.

A second evaluator helped most on questions it had solved correctly (Figure~\ref{fig:ce}C). Where it had answered the item correctly, it outperformed self-evaluation in AUROC by $+0.083$ [0.066, 0.100] on GPT's answers and $+0.070$ [0.058, 0.082] on Gemini's; where it had also erred, the gain vanished or reversed ($-0.007$ [$-0.028$, 0.013] and $-0.113$ [$-0.136$, $-0.089$]). Cross-evaluation also helps select between the two answers. On the 1{,}492 items that exactly one model solved, rating each answer by the other model alone picked the correct one 75.7\% of the time, against 67.2\% for self-ratings (Appendix~\ref{app:crossed-selection}). When both models gave the same wrong answer, however, only 6.4\% of GPT's and 5.4\% of Gemini's versions were rated below 50\% by either evaluator (Appendix~\ref{app:crossed-shared}). This low flagging rate persists when MMLU-Pro is omitted (Appendix~\ref{app:label-sensitivity}). Shared errors are rarely flagged and leave no correct alternative among the two candidates.


\section{Related Work}
\label{sec:related}

\paragraph{Confidence in language models.} Language models can learn to predict whether they will answer a question correctly and can judge whether a given answer is true \citep{kadavath2022language}; they can also state their confidence in words \citep{lin2022teaching,tian2023just,xiong2024can} or reveal uncertainty through the variability of repeated samples \citep{kuhn2023semantic,farquhar2024semantic}. Our three reports are verbal versions of these question- and answer-level judgments, elicited on one scale and compared on the same answers across 15 benchmarks.

\paragraph{Solving vs. judging.} Models can produce outputs they do not fully understand \citep{west2024paradox}, often fail to locate errors in chains of reasoning \citep{tyen2024errors}, and rarely improve by correcting themselves without external feedback \citep{huang2024cannot,kamoi2024survey}; GPT-4 reported full confidence on every GSM8K answer while answering 93.6\% correctly \citep{xiong2024can}. Section~\ref{sec:solving-is-not-knowing} shows that high accuracy does not guarantee that confidence ranks errors well, and that a review can change the judgment of an answer without changing the answer. Holding the answer fixed thus separates recognizing an error from repairing it, the two steps that self-correction combines.

\paragraph{Difficulty.} Verification is most reliable on easy problems and generally tracks the verifier's own problem-solving ability \citep{zhou2026variation}; many strong LLM judges barely beat chance when choosing between responses to challenging questions \citep{tan2025judgebench}. Section~\ref{sec:difficulty} finds a similar difficulty gradient when models judge their own answers: discrimination is weaker on questions missed by a reference model, and review offers limited improvement on these questions. A large share of pooled AUROC comes from pairs that cross difficulty tiers, which a question-only forecast such as \pik\ \citep{kadavath2022language} already ranks well. In the terms of diagnostic testing, difficulty is a covariate that shifts both the marker and the outcome, and discrimination should be assessed within its strata \citep{janes2008adjusting}.

\paragraph{Models evaluating models.} LLM judges favor their own outputs \citep{panickssery2024recognize,liu2024narcissistic,wataoka2024selfpreference}, but how much of this preference reflects authorship rather than answer quality or general judging errors is debated \citep{chen2025beyond,roytburg2026narcissists,pombal2026rubric}. Judges also favor models whose mistakes resemble their own \citep{goel2025thinkalike} and, without a verified human-written reference, agree with experts mainly on questions they can answer \citep{krumdick2025nofree}; models share many errors \citep{kim2025correlated}, and disagreement between models can expose confident errors \citep{hamidieh2026complementing}. Section~\ref{sec:crossed} shows how these patterns combine in individual answers. Evaluators rate matching answers highly, with little self-source advantage when final answers coincide. Cross-evaluation improves discrimination most when the evaluator answered the question correctly, while shared errors are rarely flagged and offer no correct candidate to select. Extending the label-free logical-consistency perspective of \citet{fluri2024evaluating}, the same section derives a bound on mean overconfidence from ratings of mutually exclusive answers.

\section{Conclusion \& Discussion}
\label{sec:discussion}

The exam strategy we began with rests on two judgments: which question to attempt and which answer to check. Our models' question-only forecasts capture useful differences in difficulty. Recognizing errors within hard questions is more demanding: confidence is less discriminating there, and review yields limited improvement. Peer evaluation is most useful on questions the evaluator has solved correctly. Shared errors can survive both judgments.

Metacognition research has long cautioned that apparent self-knowledge can arise without evaluating the choice itself. An animal can selectively decline difficult trials simply by sensing stimulus strength \citep{fleming2026integrative}. Here, much of pooled discrimination comes from comparisons across difficulty tiers, and question-only forecasts already rank these pairs well. When models evaluate completed answers, their judgments resemble the consensus patterns of human confidence \citep{koriat2008consensuality,koriat2012self}: evaluators give higher confidence to answers that match their own, and shared errors pass through peer oversight largely unflagged.

The confidence reports studied here are elicited by prompts and require no task-specific training. Training-based monitors offer complementary information: probes predict correctness from hidden states \citep{azaria2023internal,orgad2025know}, and fine-tuned models learn to predict their own accuracy or behavior \citep{kadavath2022language,kapoor2024taught,binder2025lookinginward}. Such monitors can outperform prompted reports \citep{kapoor2024taught}. Probes characterize information available in a model's representations, while prompted reports characterize its expressed judgment, a distinction also made between neural decoding and metacognitive reports in cognitive science \citep{fleming2026integrative}. Discrimination within difficulty strata offers a common test of their usefulness for detecting errors.

Holding answers fixed and crossing solvers with evaluators provides a way to test which monitoring signals retain their value on hard questions and shared errors. Closing the loop from monitoring to control requires determining when an unconfident agent should pause, when it should seek peer verification, and when it must defer to a stronger model entirely---decisions that determine whether compound errors cascade or self-correct in multi-step agentic execution.

These limits are especially consequential for autonomous scientific exploration, where models generate novel hypotheses and complex analyses that no pre-existing verifier can trivially grade. Language models also offer tractable systems for studying metacognition: the information supplied to an evaluator can be controlled while the candidate answer is held fixed. Whether their human-like confidence patterns are inherited from human text or arise, as some confidence biases do in far simpler networks \citep{webb2023natural}, from how a system maps evidence to confidence remains an open question. Back in the exam room, knowing which questions are hard helps a student budget their time; knowing which answers are wrong guides their corrections. For the models studied here, a second look helps unevenly, and a second model helps most on questions it has answered correctly. Shared mistakes remain the harder test.

\section*{AI use statement}
In this work, we used generative AI tools to implement methods, provide methodological feedback, assist with translation, clean and reformat datasets, support qualitative and thematic data analysis, and interpret results. They also assisted with the design of robustness analyses, including held-out prediction and blinded answer-key checks. We have not used generative AI tools to propose or refine hypotheses. Generating synthetic datasets, developing theoretical models or conceptual frameworks, formulating mathematical claims, and writing or supplying critical ingredients for mathematical proofs are not applicable to this work. Additionally, we used generative AI tools to create or modify scientific figures or images, identify relevant literature, edit the paper to improve readability, brainstorm, search for information, create or edit software code, and summarize or analyse existing literature. We have reviewed all AI-assisted work. We checked example answers and judgements by the LLMs. We take responsibility for the final content of this work, including text, claims or artifacts produced with the aid of generative AI.

\section*{Ethics Statement}
This work involves no human participants or personal data; all experiments use publicly available benchmarks and models under their respective licenses and terms of use. To avoid redistributing benchmark content or leaking test items into future training data, the item records we will release omit benchmark text, images, and raw model responses. The study introduces no new capabilities. Its findings bear on the oversight of AI systems: they caution against relying on a model's confidence, or on another model's judgment, to catch errors on hard questions or errors that models share. The authors declare no competing interests.

\section*{Reproducibility Statement}
Section~\ref{sec:evaluation} defines the three confidence conditions and the items on which they are compared, and Section~\ref{sec:metacognitive-monitoring-atlas} defines within-task AUROC. Appendix~\ref{app:confidence-prompts} specifies the elicitation protocol through an information-flow diagram, pseudocode (Appendix~\ref{app:confidence-pseudocode}), verbatim prompt templates (Appendices~\mbox{\ref{app:prompt-pre-answer}--\ref{app:prompt-post-answer}}), and request construction, decoding, and model-specific settings (Appendix~\ref{app:confidence-implementation}). Appendix~\ref{app:protocol} lists the benchmarks, item counts, and graders (Table~\ref{tab:benchmarks}), describes how the mathematical grades were verified, and gives the treatment of abstentions, format failures, and missing confidence values; and Appendix~\ref{app:self-evaluation-results} reports accuracy on every benchmark and the coverage of valid confidence reports. Appendix~\ref{app:difficulty-robustness} gives the robustness checks of the difficulty analysis, and Appendices~\ref{app:crossed-confidence-prompts} and~\ref{app:crossed-population} give the source-neutral prompt, population, and estimators of the crossed evaluation; the construction of every interval is stated with the result it accompanies. Appendix~\ref{app:additional-analyses} provides stratified retention and format recovery, recorded generation parameters, the MATH-500 content inspection, held-out incremental prediction, and label-sensitivity analyses. 

\bibliographystyle{iclr2027_conference}
\bibliography{references}

\begin{thebibliography}{57}
\providecommand{\natexlab}[1]{#1}
\providecommand{\url}[1]{\texttt{#1}}
\expandafter\ifx\csname urlstyle\endcsname\relax
  \providecommand{\doi}[1]{doi: #1}\else
  \providecommand{\doi}{doi: \begingroup \urlstyle{rm}\Url}\fi

\bibitem[Ackerman \& Thompson(2017)Ackerman and
  Thompson]{ackerman2017metareasoning}
Rakefet Ackerman and Valerie~A. Thompson.
\newblock Meta-reasoning: Monitoring and control of thinking and reasoning.
\newblock \emph{Trends in Cognitive Sciences}, 21\penalty0 (8):\penalty0
  607--617, 2017.
\newblock \doi{10.1016/j.tics.2017.05.004}.

\bibitem[Azaria \& Mitchell(2023)Azaria and Mitchell]{azaria2023internal}
Amos Azaria and Tom Mitchell.
\newblock The internal state of an {LLM} knows when it's lying.
\newblock In \emph{Findings of the Association for Computational Linguistics:
  EMNLP 2023}, pp.\  967--976, 2023.
\newblock \doi{10.18653/v1/2023.findings-emnlp.68}.
\newblock URL \url{https://aclanthology.org/2023.findings-emnlp.68/}.

\bibitem[Bai et~al.(2025)Bai, Tu, Zhang, Peng, Wang, Lv, Cao, Xu, Hou, Dong,
  Tang, and Li]{bai2024longbench2}
Yushi Bai, Shangqing Tu, Jiajie Zhang, Hao Peng, Xiaozhi Wang, Xin Lv, Shulin
  Cao, Jiazheng Xu, Lei Hou, Yuxiao Dong, Jie Tang, and Juanzi Li.
\newblock {LongBench v2}: Towards deeper understanding and reasoning on
  realistic long-context multitasks.
\newblock In \emph{Proceedings of the 63rd Annual Meeting of the Association
  for Computational Linguistics (Volume 1: Long Papers)}, pp.\  3639--3664,
  2025.
\newblock \doi{10.18653/v1/2025.acl-long.183}.
\newblock URL \url{https://aclanthology.org/2025.acl-long.183/}.

\bibitem[Binder et~al.(2025)Binder, Chua, Korbak, Sleight, Hughes, Long, Perez,
  Turpin, and Evans]{binder2025lookinginward}
Felix~Jedidja Binder, James Chua, Tomek Korbak, Henry Sleight, John Hughes,
  Robert Long, Ethan Perez, Miles Turpin, and Owain Evans.
\newblock Looking inward: Language models can learn about themselves by
  introspection.
\newblock In \emph{International Conference on Learning Representations}, 2025.
\newblock URL
  \url{https://proceedings.iclr.cc/paper_files/paper/2025/hash/0a6059857ae5c82ea9726ee9282a7145-Abstract-Conference.html}.

\bibitem[{Center for AI Safety} et~al.(2026){Center for AI Safety}, Phan,
  Gatti, Li, et~al.]{phan2025hle}
{Center for AI Safety}, Long Phan, Alice Gatti, Nathaniel Li, et~al.
\newblock A benchmark of expert-level academic questions to assess {AI}
  capabilities.
\newblock \emph{Nature}, 649\penalty0 (8099):\penalty0 1139--1146, 2026.
\newblock \doi{10.1038/s41586-025-09962-4}.
\newblock URL \url{https://doi.org/10.1038/s41586-025-09962-4}.

\bibitem[Chen(2026)]{chen2026cofailure}
Josef Chen.
\newblock When does combining language models help? {A} co-failure ceiling on
  routing, voting, and mixture-of-agents across 67 frontier models.
\newblock \emph{arXiv preprint arXiv:2606.27288}, 2026.
\newblock URL \url{https://arxiv.org/abs/2606.27288}.

\bibitem[Chen et~al.(2025{\natexlab{a}})Chen, Wei, Zhu, Feng, and
  Meng]{chen2025prefer}
Wei-Lin Chen, Zhepei Wei, Xinyu Zhu, Shi Feng, and Yu~Meng.
\newblock Do {LLM} evaluators prefer themselves for a reason?
\newblock \emph{arXiv preprint arXiv:2504.03846}, 2025{\natexlab{a}}.
\newblock URL \url{https://arxiv.org/abs/2504.03846v3}.

\bibitem[Chen et~al.(2025{\natexlab{b}})Chen, Wang, Zhang, Hu, and
  Lin]{chen2025beyond}
Zhi-Yuan Chen, Hao Wang, Xinyu Zhang, Enrui Hu, and Yankai Lin.
\newblock Beyond the surface: Measuring self-preference in {LLM} judgments.
\newblock In \emph{Proceedings of the 2025 Conference on Empirical Methods in
  Natural Language Processing}, pp.\  1653--1672, 2025{\natexlab{b}}.
\newblock \doi{10.18653/v1/2025.emnlp-main.86}.
\newblock URL \url{https://aclanthology.org/2025.emnlp-main.86/}.

\bibitem[Farquhar et~al.(2024)Farquhar, Kossen, Kuhn, and
  Gal]{farquhar2024semantic}
Sebastian Farquhar, Jannik Kossen, Lorenz Kuhn, and Yarin Gal.
\newblock Detecting hallucinations in large language models using semantic
  entropy.
\newblock \emph{Nature}, 630\penalty0 (8017):\penalty0 625--630, 2024.
\newblock \doi{10.1038/s41586-024-07421-0}.
\newblock URL \url{https://www.nature.com/articles/s41586-024-07421-0}.

\bibitem[Fleming(2026)]{fleming2026integrative}
Stephen~M. Fleming.
\newblock Towards an integrative neuroscience of metacognition.
\newblock \emph{Nature Reviews Neuroscience}, 2026.
\newblock \doi{10.1038/s41583-026-01081-x}.

\bibitem[Fleming \& Lau(2014)Fleming and Lau]{fleming2014measure}
Stephen~M. Fleming and Hakwan~C. Lau.
\newblock How to measure metacognition.
\newblock \emph{Frontiers in Human Neuroscience}, 8:\penalty0 443, 2014.
\newblock \doi{10.3389/fnhum.2014.00443}.
\newblock URL
  \url{https://www.frontiersin.org/journals/human-neuroscience/articles/10.3389/fnhum.2014.00443/full}.

\bibitem[Fluri et~al.(2024)Fluri, Paleka, and Tram{\`e}r]{fluri2024evaluating}
Lukas Fluri, Daniel Paleka, and Florian Tram{\`e}r.
\newblock Evaluating superhuman models with consistency checks.
\newblock In \emph{IEEE Conference on Secure and Trustworthy Machine Learning
  (SaTML)}, pp.\  194--232, 2024.
\newblock \doi{10.1109/SaTML59370.2024.00017}.
\newblock URL \url{https://doi.org/10.1109/SaTML59370.2024.00017}.

\bibitem[Garg \& Sagtani(2026)Garg and Sagtani]{garg2026unsolvability}
Saloni Garg and Amit Sagtani.
\newblock Unsolvability ceiling in multi-{LLM} routing: An empirical study of
  evaluation artifacts.
\newblock \emph{arXiv preprint arXiv:2605.07395}, 2026.
\newblock URL \url{https://arxiv.org/abs/2605.07395}.

\bibitem[Goel et~al.(2025)Goel, Str{\"u}ber, Auzina, Chandra, Kumaraguru,
  Kiela, Prabhu, Bethge, and Geiping]{goel2025thinkalike}
Shashwat Goel, Joschka Str{\"u}ber, Ilze~Amanda Auzina, Karuna~K. Chandra,
  Ponnurangam Kumaraguru, Douwe Kiela, Ameya Prabhu, Matthias Bethge, and Jonas
  Geiping.
\newblock Great models think alike and this undermines {AI} oversight.
\newblock In \emph{Proceedings of the 42nd International Conference on Machine
  Learning}, volume 267 of \emph{Proceedings of Machine Learning Research},
  pp.\  19621--19678, 2025.
\newblock URL \url{https://proceedings.mlr.press/v267/goel25b.html}.

\bibitem[Hamidieh et~al.(2026)Hamidieh, Thost, Gerych, Yurochkin, and
  Ghassemi]{hamidieh2026complementing}
Kimia Hamidieh, Veronika Thost, Walter Gerych, Mikhail Yurochkin, and Marzyeh
  Ghassemi.
\newblock Complementing self-consistency with cross-model disagreement for
  uncertainty quantification.
\newblock In \emph{International Conference on Learning Representations}, 2026.
\newblock URL \url{https://iclr.cc/virtual/2026/poster/10007682}.

\bibitem[He et~al.(2024)He, Luo, Bai, Hu, Thai, Shen, Hu, Han, Huang, Zhang,
  Liu, Qi, Liu, and Sun]{he2024olympiadbench}
Chaoqun He, Renjie Luo, Yuzhuo Bai, Shengding Hu, Zhen Thai, Junhao Shen, Jinyi
  Hu, Xu~Han, Yujie Huang, Yuxiang Zhang, Jie Liu, Lei Qi, Zhiyuan Liu, and
  Maosong Sun.
\newblock {OlympiadBench}: A challenging benchmark for promoting {AGI} with
  olympiad-level bilingual multimodal scientific problems.
\newblock In \emph{Proceedings of the 62nd Annual Meeting of the Association
  for Computational Linguistics (Volume 1: Long Papers)}, pp.\  3828--3850,
  2024.
\newblock \doi{10.18653/v1/2024.acl-long.211}.
\newblock URL \url{https://aclanthology.org/2024.acl-long.211/}.

\bibitem[Hendrycks et~al.(2021)Hendrycks, Burns, Kadavath, Arora, Basart, Tang,
  Song, and Steinhardt]{hendrycks2021math}
Dan Hendrycks, Collin Burns, Saurav Kadavath, Akul Arora, Steven Basart, Eric
  Tang, Dawn Song, and Jacob Steinhardt.
\newblock Measuring mathematical problem solving with the {MATH} dataset.
\newblock In \emph{Proceedings of the Neural Information Processing Systems
  Track on Datasets and Benchmarks}, volume~1, 2021.
\newblock URL
  \url{https://datasets-benchmarks-proceedings.neurips.cc/paper/2021/hash/be83ab3ecd0db773eb2dc1b0a17836a1-Abstract-round2.html}.

\bibitem[Huang et~al.(2024)Huang, Chen, Mishra, Zheng, Yu, Song, and
  Zhou]{huang2024cannot}
Jie Huang, Xinyun Chen, Swaroop Mishra, Huaixiu~Steven Zheng, Adams~Wei Yu,
  Xinying Song, and Denny Zhou.
\newblock Large language models cannot self-correct reasoning yet.
\newblock In \emph{International Conference on Learning Representations},
  volume 2024, pp.\  32808--32824, 2024.
\newblock URL
  \url{https://proceedings.iclr.cc/paper_files/paper/2024/file/8b4add8b0aa8749d80a34ca5d941c355-Paper-Conference.pdf}.

\bibitem[Jain et~al.(2025)Jain, Han, Gu, Li, Yan, Zhang, Wang, Solar-Lezama,
  Sen, and Stoica]{jain2024livecodebench}
Naman Jain, King Han, Alex Gu, Wen-Ding Li, Fanjia Yan, Tianjun Zhang, Sida
  Wang, Armando Solar-Lezama, Koushik Sen, and Ion Stoica.
\newblock {LiveCodeBench}: Holistic and contamination free evaluation of large
  language models for code.
\newblock In \emph{International Conference on Learning Representations},
  volume 2025, pp.\  58791--58831, 2025.
\newblock URL
  \url{https://proceedings.iclr.cc/paper_files/paper/2025/file/94074dd5a072d28ff75a76dabed43767-Paper-Conference.pdf}.

\bibitem[Janes \& Pepe(2008)Janes and Pepe]{janes2008adjusting}
Holly Janes and Margaret~S. Pepe.
\newblock Adjusting for covariates in studies of diagnostic, screening, or
  prognostic markers: An old concept in a new setting.
\newblock \emph{American Journal of Epidemiology}, 168\penalty0 (1):\penalty0
  89--97, 2008.
\newblock \doi{10.1093/aje/kwn099}.
\newblock URL \url{https://doi.org/10.1093/aje/kwn099}.

\bibitem[Kadavath et~al.(2022)Kadavath, Conerly, Askell, Henighan, Drain,
  Perez, Schiefer, Hatfield-Dodds, DasSarma, Tran-Johnson, Johnston, El-Showk,
  Jones, Elhage, Hume, Chen, Bai, Bowman, Fort, Ganguli, Hernandez, Jacobson,
  Kernion, Kravec, Lovitt, Ndousse, Olsson, Ringer, Amodei, Brown, Clark,
  Joseph, Mann, McCandlish, Olah, and Kaplan]{kadavath2022language}
Saurav Kadavath, Tom Conerly, Amanda Askell, Tom Henighan, Dawn Drain, Ethan
  Perez, Nicholas Schiefer, Zac Hatfield-Dodds, Nova DasSarma, Eli
  Tran-Johnson, Scott Johnston, Sheer El-Showk, Andy Jones, Nelson Elhage,
  Tristan Hume, Anna Chen, Yuntao Bai, Sam Bowman, Stanislav Fort, Deep
  Ganguli, Danny Hernandez, Josh Jacobson, Jackson Kernion, Shauna Kravec,
  Liane Lovitt, Kamal Ndousse, Catherine Olsson, Sam Ringer, Dario Amodei, Tom
  Brown, Jack Clark, Nicholas Joseph, Ben Mann, Sam McCandlish, Chris Olah, and
  Jared Kaplan.
\newblock Language models (mostly) know what they know.
\newblock \emph{arXiv preprint arXiv:2207.05221}, 2022.
\newblock URL \url{https://arxiv.org/abs/2207.05221}.

\bibitem[Kamoi et~al.(2024)Kamoi, Zhang, Zhang, Han, and
  Zhang]{kamoi2024survey}
Ryo Kamoi, Yusen Zhang, Nan Zhang, Jiawei Han, and Rui Zhang.
\newblock When can {LLMs} actually correct their own mistakes? {A} critical
  survey of self-correction of {LLMs}.
\newblock \emph{Transactions of the Association for Computational Linguistics},
  12:\penalty0 1417--1440, 2024.
\newblock \doi{10.1162/tacl_a_00713}.

\bibitem[Kapoor et~al.(2024)Kapoor, Gruver, Roberts, Collins, Pal, Bhatt,
  Weller, Dooley, Goldblum, and Wilson]{kapoor2024taught}
Sanyam Kapoor, Nate Gruver, Manley Roberts, Katherine Collins, Arka Pal, Umang
  Bhatt, Adrian Weller, Samuel Dooley, Micah Goldblum, and Andrew~Gordon
  Wilson.
\newblock Large language models must be taught to know what they don't know.
\newblock In \emph{Advances in Neural Information Processing Systems},
  volume~37, 2024.
\newblock URL \url{https://arxiv.org/abs/2406.08391}.

\bibitem[Kim et~al.(2025)Kim, Garg, Peng, and Garg]{kim2025correlated}
Elliot~Myunghoon Kim, Avi Garg, Kenny Peng, and Nikhil Garg.
\newblock Correlated errors in large language models.
\newblock In \emph{Proceedings of the 42nd International Conference on Machine
  Learning}, volume 267 of \emph{Proceedings of Machine Learning Research},
  pp.\  30038--30066, 2025.
\newblock URL \url{https://proceedings.mlr.press/v267/kim25e.html}.

\bibitem[Koriat(2008)]{koriat2008consensuality}
Asher Koriat.
\newblock Subjective confidence in one's answers: The consensuality principle.
\newblock \emph{Journal of Experimental Psychology: Learning, Memory, and
  Cognition}, 34\penalty0 (4):\penalty0 945--959, 2008.
\newblock \doi{10.1037/0278-7393.34.4.945}.

\bibitem[Koriat(2012)]{koriat2012self}
Asher Koriat.
\newblock The self-consistency model of subjective confidence.
\newblock \emph{Psychological Review}, 119\penalty0 (1):\penalty0 80--113,
  2012.
\newblock \doi{10.1037/a0025648}.

\bibitem[Kruger \& Dunning(1999)Kruger and Dunning]{kruger1999unskilled}
Justin Kruger and David Dunning.
\newblock Unskilled and unaware of it: How difficulties in recognizing one's
  own incompetence lead to inflated self-assessments.
\newblock \emph{Journal of Personality and Social Psychology}, 77\penalty0
  (6):\penalty0 1121--1134, 1999.
\newblock \doi{10.1037/0022-3514.77.6.1121}.

\bibitem[Krumdick et~al.(2025)Krumdick, Lovering, Reddy, Ebner, and
  Tanner]{krumdick2025nofree}
Michael Krumdick, Charles Lovering, Varshini Reddy, Seth Ebner, and Chris
  Tanner.
\newblock No free labels: Limitations of {LLM}-as-a-judge without human
  grounding.
\newblock \emph{arXiv preprint arXiv:2503.05061}, 2025.
\newblock URL \url{https://arxiv.org/abs/2503.05061}.

\bibitem[Kuhn et~al.(2023)Kuhn, Gal, and Farquhar]{kuhn2023semantic}
Lorenz Kuhn, Yarin Gal, and Sebastian Farquhar.
\newblock Semantic uncertainty: Linguistic invariances for uncertainty
  estimation in natural language generation.
\newblock In \emph{International Conference on Learning Representations}, 2023.
\newblock URL \url{https://arxiv.org/abs/2302.09664}.

\bibitem[Kydl{\'i}{\v{c}}ek(2025)]{kydlicek2025mathverify}
Hynek Kydl{\'i}{\v{c}}ek.
\newblock {Math-Verify}: Math verification library, 2025.
\newblock URL \url{https://github.com/huggingface/Math-Verify}.

\bibitem[Lichtenstein \& Fischhoff(1977)Lichtenstein and
  Fischhoff]{lichtenstein1977know}
Sarah Lichtenstein and Baruch Fischhoff.
\newblock Do those who know more also know more about how much they know?
\newblock \emph{Organizational Behavior and Human Performance}, 20\penalty0
  (2):\penalty0 159--183, 1977.
\newblock \doi{10.1016/0030-5073(77)90001-0}.

\bibitem[Lightman et~al.(2024)Lightman, Kosaraju, Burda, Edwards, Baker, Lee,
  Leike, Schulman, Sutskever, and Cobbe]{lightman2023verify}
Hunter Lightman, Vineet Kosaraju, Yura Burda, Harri Edwards, Bowen Baker, Teddy
  Lee, Jan Leike, John Schulman, Ilya Sutskever, and Karl Cobbe.
\newblock Let's verify step by step.
\newblock In \emph{International Conference on Learning Representations},
  volume 2024, pp.\  39578--39601, 2024.
\newblock URL
  \url{https://proceedings.iclr.cc/paper_files/paper/2024/file/aca97732e30bcf1303bc22ac3924fd16-Paper-Conference.pdf}.

\bibitem[Lin et~al.(2025)Lin, Le~Bras, Richardson, Sabharwal, Poovendran,
  Clark, and Choi]{lin2024zebralogic}
Bill~Yuchen Lin, Ronan Le~Bras, Kyle Richardson, Ashish Sabharwal, Radha
  Poovendran, Peter Clark, and Yejin Choi.
\newblock {ZebraLogic}: On the scaling limits of {LLM}s for logical reasoning.
\newblock In \emph{Proceedings of the 42nd International Conference on Machine
  Learning}, volume 267 of \emph{Proceedings of Machine Learning Research},
  pp.\  37889--37905, 2025.
\newblock URL \url{https://proceedings.mlr.press/v267/lin25i.html}.

\bibitem[Lin et~al.(2022)Lin, Hilton, and Evans]{lin2022teaching}
Stephanie Lin, Jacob Hilton, and Owain Evans.
\newblock Teaching models to express their uncertainty in words.
\newblock \emph{Transactions on Machine Learning Research}, 2022.
\newblock URL \url{https://openreview.net/forum?id=8s8K2UZGTZ}.

\bibitem[Liu et~al.(2025)Liu, Liu, Xiao, Wang, Liu, Gao, Zhang, Zhang, and
  Chen]{chen2024livemathbench}
Junnan Liu, Hongwei Liu, Linchen Xiao, Ziyi Wang, Kuikun Liu, Songyang Gao,
  Wenwei Zhang, Songyang Zhang, and Kai Chen.
\newblock Are your {LLM}s capable of stable reasoning?
\newblock In \emph{Findings of the Association for Computational Linguistics:
  ACL 2025}, pp.\  17594--17632, 2025.
\newblock \doi{10.18653/v1/2025.findings-acl.905}.
\newblock URL \url{https://aclanthology.org/2025.findings-acl.905/}.

\bibitem[Liu et~al.(2024)Liu, Moosavi, and Lin]{liu2024narcissistic}
Yiqi Liu, Nafise~Sadat Moosavi, and Chenghua Lin.
\newblock {LLMs} as narcissistic evaluators: When ego inflates evaluation
  scores.
\newblock In \emph{Findings of the Association for Computational Linguistics:
  ACL 2024}, pp.\  12688--12701, 2024.
\newblock \doi{10.18653/v1/2024.findings-acl.753}.
\newblock URL \url{https://aclanthology.org/2024.findings-acl.753/}.

\bibitem[Nelson \& Narens(1990)Nelson and Narens]{nelson1990metamemory}
Thomas~O. Nelson and Louis Narens.
\newblock Metamemory: A theoretical framework and new findings.
\newblock In Gordon~H. Bower (ed.), \emph{The Psychology of Learning and
  Motivation}, volume~26, pp.\  125--173. Academic Press, 1990.
\newblock \doi{10.1016/S0079-7421(08)60053-5}.

\bibitem[{OpenAI}(2024)]{openai2024simpleqa}
{OpenAI}.
\newblock Introducing {SimpleQA}, 2024.
\newblock URL \url{https://openai.com/index/introducing-simpleqa/}.

\bibitem[Orgad et~al.(2025)Orgad, Toker, Gekhman, Reichart, Szpektor, Kotek,
  and Belinkov]{orgad2025know}
Hadas Orgad, Michael Toker, Zorik Gekhman, Roi Reichart, Idan Szpektor, Hadas
  Kotek, and Yonatan Belinkov.
\newblock {LLMs} know more than they show: On the intrinsic representation of
  {LLM} hallucinations.
\newblock In \emph{International Conference on Learning Representations},
  volume 2025, pp.\  66880--66913, 2025.
\newblock URL
  \url{https://proceedings.iclr.cc/paper_files/paper/2025/file/a712d461e57201efe35d429a6f1731c1-Paper-Conference.pdf}.

\bibitem[Panickssery et~al.(2024)Panickssery, Bowman, and
  Feng]{panickssery2024recognize}
Arjun Panickssery, Samuel~R. Bowman, and Shi Feng.
\newblock {LLM} evaluators recognize and favor their own generations.
\newblock In \emph{Advances in Neural Information Processing Systems},
  volume~37, pp.\  68772--68802. Curran Associates, Inc., 2024.
\newblock \doi{10.52202/079017-2197}.
\newblock URL
  \url{https://proceedings.neurips.cc/paper_files/paper/2024/file/7f1f0218e45f5414c79c0679633e47bc-Paper-Conference.pdf}.

\bibitem[Pombal et~al.(2026)Pombal, Rei, and Martins]{pombal2026rubric}
Jos{\'e} Pombal, Ricardo Rei, and Andr{\'e} F.~T. Martins.
\newblock Self-preference bias in rubric-based evaluation of large language
  models.
\newblock In \emph{Conference on Language Modeling}, 2026.
\newblock URL \url{https://colm.cc/virtual/2026/poster/2533}.

\bibitem[Rein et~al.(2024)Rein, Hou, Stickland, Petty, Pang, Dirani, Michael,
  and Bowman]{rein2024gpqa}
David Rein, Betty~Li Hou, Asa~Cooper Stickland, Jackson Petty, Richard~Yuanzhe
  Pang, Julien Dirani, Julian Michael, and Samuel~R. Bowman.
\newblock {GPQA}: A graduate-level google-proof {Q\&A} benchmark.
\newblock In \emph{Proceedings of the First Conference on Language Modeling},
  2024.
\newblock URL \url{https://openreview.net/forum?id=Ti67584b98}.

\bibitem[Roytburg et~al.(2026)Roytburg, Bozoukov, Nguyen, Barzdukas, Puig-Hall,
  and Oozeer]{roytburg2026narcissists}
Dani Roytburg, Matthew Bozoukov, Matthew Nguyen, Jou Barzdukas, Mackenzie
  Puig-Hall, and Narmeen Oozeer.
\newblock Are {LLM} evaluators really narcissists? sanity checking
  self-preference evaluations.
\newblock In \emph{Proceedings of the 43rd International Conference on Machine
  Learning}, volume 306 of \emph{Proceedings of Machine Learning Research},
  2026.
\newblock URL \url{https://icml.cc/virtual/2026/poster/61230}.

\bibitem[Sprague et~al.(2024)Sprague, Ye, Bostrom, Chaudhuri, and
  Durrett]{sprague2024musr}
Zayne Sprague, Xi~Ye, Kaj Bostrom, Swarat Chaudhuri, and Greg Durrett.
\newblock {MuSR}: Testing the limits of chain-of-thought with multistep soft
  reasoning.
\newblock In \emph{International Conference on Learning Representations},
  volume 2024, pp.\  14670--14728, 2024.
\newblock URL
  \url{https://proceedings.iclr.cc/paper_files/paper/2024/file/3f8c7eb848ffec848f3ed2b7ca44915d-Paper-Conference.pdf}.

\bibitem[Srivastava et~al.(2023)]{srivastava2022bigbench}
Aarohi Srivastava et~al.
\newblock Beyond the imitation game: Quantifying and extrapolating the
  capabilities of language models.
\newblock \emph{Transactions on Machine Learning Research}, 2023.
\newblock URL \url{https://openreview.net/forum?id=uyTL5Bvosj}.

\bibitem[Tan et~al.(2025)Tan, Zhuang, Montgomery, Tang, Cuadron, Wang, Popa,
  and Stoica]{tan2025judgebench}
Sijun Tan, Siyuan Zhuang, Kyle Montgomery, William~Y. Tang, Alejandro Cuadron,
  Chenguang Wang, Raluca~Ada Popa, and Ion Stoica.
\newblock {JudgeBench}: A benchmark for evaluating {LLM}-based judges.
\newblock In \emph{International Conference on Learning Representations},
  volume 2025, pp.\  63277--63303, 2025.
\newblock URL
  \url{https://proceedings.iclr.cc/paper_files/paper/2025/file/9e720fce64f91114c49cfd640d821da3-Paper-Conference.pdf}.

\bibitem[Tian et~al.(2023)Tian, Mitchell, Zhou, Sharma, Rafailov, Yao, Finn,
  and Manning]{tian2023just}
Katherine Tian, Eric Mitchell, Allan Zhou, Archit Sharma, Rafael Rafailov,
  Huaxiu Yao, Chelsea Finn, and Christopher~D. Manning.
\newblock Just ask for calibration: Strategies for eliciting calibrated
  confidence scores from language models fine-tuned with human feedback.
\newblock In \emph{Proceedings of the 2023 Conference on Empirical Methods in
  Natural Language Processing}, pp.\  5433--5442, 2023.
\newblock \doi{10.18653/v1/2023.emnlp-main.330}.
\newblock URL \url{https://aclanthology.org/2023.emnlp-main.330/}.

\bibitem[Tyen et~al.(2024)Tyen, Mansoor, Carbune, Chen, and
  Mak]{tyen2024errors}
Gladys Tyen, Hassan Mansoor, Victor Carbune, Peter Chen, and Tony Mak.
\newblock {LLMs} cannot find reasoning errors, but can correct them given the
  error location.
\newblock In \emph{Findings of the Association for Computational Linguistics:
  ACL 2024}, pp.\  13894--13908, 2024.
\newblock \doi{10.18653/v1/2024.findings-acl.826}.
\newblock URL \url{https://aclanthology.org/2024.findings-acl.826/}.

\bibitem[Wang et~al.(2024{\natexlab{a}})Wang, Pan, Shi, Lu, Ren, Zhou, Zhan,
  and Li]{wang2024mathvision}
Ke~Wang, Junting Pan, Weikang Shi, Zimu Lu, Houxing Ren, Aojun Zhou, Mingjie
  Zhan, and Hongsheng Li.
\newblock Measuring multimodal mathematical reasoning with the {MATH-Vision}
  dataset.
\newblock In \emph{Advances in Neural Information Processing Systems},
  volume~37, pp.\  95095--95169. Curran Associates, Inc., 2024{\natexlab{a}}.
\newblock \doi{10.52202/079017-3014}.
\newblock URL
  \url{https://proceedings.neurips.cc/paper_files/paper/2024/file/ad0edc7d5fa1a783f063646968b7315b-Paper-Datasets_and_Benchmarks_Track.pdf}.

\bibitem[Wang et~al.(2024{\natexlab{b}})Wang, Ma, Zhang, Ni, Chandra, Guo, Ren,
  Arulraj, He, Jiang, Li, Ku, Wang, Zhuang, Fan, Yue, and
  Chen]{wang2024mmlupro}
Yubo Wang, Xueguang Ma, Ge~Zhang, Yuansheng Ni, Abhranil Chandra, Shiguang Guo,
  Weiming Ren, Aaran Arulraj, Xuan He, Ziyan Jiang, Tianle Li, Max Ku, Kai
  Wang, Alex Zhuang, Rongqi Fan, Xiang Yue, and Wenhu Chen.
\newblock {MMLU-Pro}: A more robust and challenging multi-task language
  understanding benchmark.
\newblock In \emph{Advances in Neural Information Processing Systems},
  volume~37, pp.\  95266--95290. Curran Associates, Inc., 2024{\natexlab{b}}.
\newblock \doi{10.52202/079017-3018}.
\newblock URL
  \url{https://proceedings.neurips.cc/paper_files/paper/2024/file/ad236edc564f3e3156e1b2feafb99a24-Paper-Datasets_and_Benchmarks_Track.pdf}.

\bibitem[Wataoka et~al.(2024)Wataoka, Takahashi, and
  Ri]{wataoka2024selfpreference}
Koki Wataoka, Tsubasa Takahashi, and Ryokan Ri.
\newblock Self-preference bias in {LLM}-as-a-judge.
\newblock \emph{arXiv preprint arXiv:2410.21819}, 2024.
\newblock URL \url{https://arxiv.org/abs/2410.21819v2}.
\newblock Version 2, revised June 2025.

\bibitem[Webb et~al.(2023)Webb, Miyoshi, So, Rajananda, and
  Lau]{webb2023natural}
Taylor~W. Webb, Kiyofumi Miyoshi, Tsz~Yan So, Sivananda Rajananda, and Hakwan
  Lau.
\newblock Natural statistics support a rational account of confidence biases.
\newblock \emph{Nature Communications}, 14\penalty0 (1):\penalty0 3992, 2023.
\newblock \doi{10.1038/s41467-023-39737-2}.
\newblock URL \url{https://doi.org/10.1038/s41467-023-39737-2}.

\bibitem[West et~al.(2024)West, Lu, Dziri, Brahman, Li, Hwang, Jiang, Fisher,
  Ravichander, Chandu, Newman, Koh, Ettinger, and Choi]{west2024paradox}
Peter West, Ximing Lu, Nouha Dziri, Faeze Brahman, Linjie Li, Jena~D. Hwang,
  Liwei Jiang, Jillian Fisher, Abhilasha Ravichander, Khyathi Chandu, Benjamin
  Newman, Pang~Wei Koh, Allyson Ettinger, and Yejin Choi.
\newblock The generative {AI} paradox: ``what it can create, it may not
  understand''.
\newblock In \emph{International Conference on Learning Representations}, 2024.
\newblock URL
  \url{https://proceedings.iclr.cc/paper_files/paper/2024/hash/ce208d95d020b023cba9e64031db2584-Abstract-Conference.html}.

\bibitem[White et~al.(2025)White, Dooley, Roberts, Pal, Feuer, Jain,
  Shwartz-Ziv, Jain, Saifullah, Dey, Shubh-Agrawal, Sandha, Naidu, Hegde,
  LeCun, Goldstein, Neiswanger, and Goldblum]{white2024livebench}
Colin White, Samuel Dooley, Manley Roberts, Arka Pal, Benjamin Feuer,
  Siddhartha Jain, Ravid Shwartz-Ziv, Neel Jain, Khalid Saifullah, Sreemanti
  Dey, Shubh-Agrawal, Sandeep Sandha, Siddartha Naidu, Chinmay Hegde, Yann
  LeCun, Tom Goldstein, Willie Neiswanger, and Micah Goldblum.
\newblock {LiveBench}: A challenging, contamination-limited {LLM} benchmark.
\newblock In \emph{International Conference on Learning Representations},
  volume 2025, pp.\  91595--91631, 2025.
\newblock URL
  \url{https://proceedings.iclr.cc/paper_files/paper/2025/file/e4a46394ba5378b3f9a186a5b4c650d1-Paper-Conference.pdf}.

\bibitem[Xiong et~al.(2024)Xiong, Hu, Lu, Li, Fu, He, and Hooi]{xiong2024can}
Miao Xiong, Zhiyuan Hu, Xinyang Lu, Yifei Li, Jie Fu, Junxian He, and Bryan
  Hooi.
\newblock Can {LLMs} express their uncertainty? an empirical evaluation of
  confidence elicitation in {LLMs}.
\newblock In \emph{International Conference on Learning Representations},
  volume 2024, pp.\  23650--23678, 2024.
\newblock URL
  \url{https://proceedings.iclr.cc/paper_files/paper/2024/file/6733cf15e10e2cd1d59af033c3bb8507-Paper-Conference.pdf}.

\bibitem[Yue et~al.(2025)Yue, Zheng, Ni, Wang, Zhang, Tong, Sun, Yu, Zhang,
  Sun, Su, Chen, and Neubig]{yue2024mmmupro}
Xiang Yue, Tianyu Zheng, Yuansheng Ni, Yubo Wang, Kai Zhang, Shengbang Tong,
  Yuxuan Sun, Botao Yu, Ge~Zhang, Huan Sun, Yu~Su, Wenhu Chen, and Graham
  Neubig.
\newblock {MMMU}-pro: A more robust multi-discipline multimodal understanding
  benchmark.
\newblock In \emph{Proceedings of the 63rd Annual Meeting of the Association
  for Computational Linguistics (Volume 1: Long Papers)}, pp.\  15134--15186,
  2025.
\newblock \doi{10.18653/v1/2025.acl-long.736}.
\newblock URL \url{https://aclanthology.org/2025.acl-long.736/}.

\bibitem[Zhou et~al.(2026)Zhou, Xu, Zhou, Singh, Gui, and
  Joty]{zhou2026variation}
Yefan Zhou, Austin Xu, Yilun Zhou, Janvijay Singh, Jiang Gui, and Shafiq Joty.
\newblock Variation in verification: Understanding verification dynamics in
  large language models.
\newblock In \emph{International Conference on Learning Representations}, 2026.
\newblock URL
  \url{https://proceedings.iclr.cc/paper_files/paper/2026/hash/ea21628f0fc0de542373be4c88343478-Abstract-Conference.html}.

\end{thebibliography}

\clearpage
\appendix
\section{Confidence Elicitation Protocol and Prompts}
\label{app:confidence-prompts}

\noindent\begin{minipage}{\linewidth}
  \centering
  \includegraphics[width=5.5in]{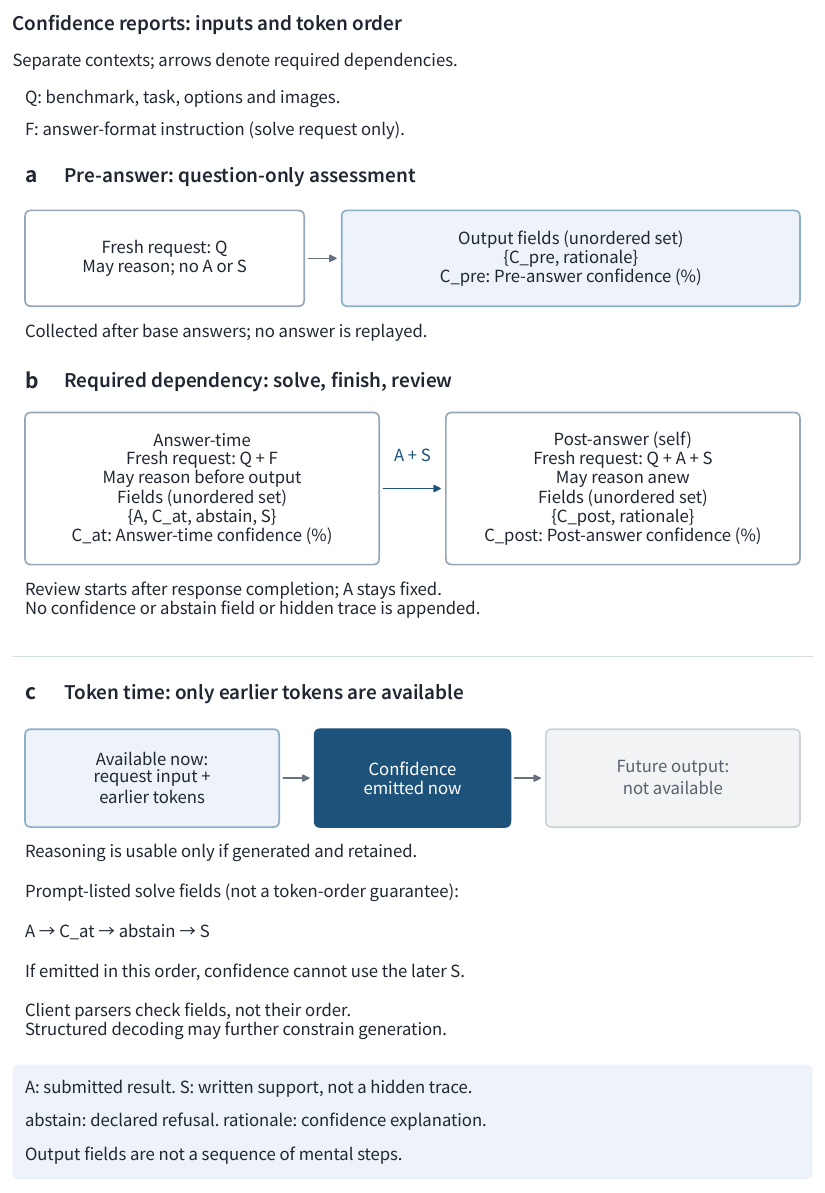}
  \captionsetup{hypcap=false}
  \captionof{figure}{\textbf{Request-level inputs and token-level chronology.} $Q$ is the task input, $F$ the solve-only answer-format instruction, $A$ the submitted answer, and $S$ its visible written solution, not hidden reasoning. $C_{\mathrm{pre}}$, $C_{\mathrm{at}}$, and $C_{\mathrm{post}}$ denote Pre-answer, Answer-time, and Post-answer confidence (0--100\%). Braces denote unordered fields; arrows indicate required dependencies. Pseudocode and model settings: Appendices~\ref{app:confidence-pseudocode} and~\ref{app:confidence-implementation}.}
  \label{fig:confidence-information-flow}
\end{minipage}
\clearpage

\subsection{Final answers and written solutions}
\label{app:answer-solution-examples}

The \texttt{answer} field contains the submitted result, such as a number, an option label, or executable code. The \texttt{solution} field contains the written explanation supporting that result.

\paragraph{Illustrative example: elementary algebra.}
For $2x+1=7$, an illustrative response would distinguish the two fields as follows:
\begin{quote}\small
\textbf{\texttt{answer}:} \texttt{3}\\
\textbf{\texttt{solution}:} Subtract 1 from both sides to obtain $2x=6$, then divide by 2 to obtain $x=3$.
\end{quote}

The following three examples are recorded GPT-5.6 Sol responses. Task descriptions are condensed; the \texttt{answer} and \texttt{solution} contents are reproduced without abridgment, with only typesetting changes.

\paragraph{Recorded mathematics: MATH-500.}
Task: find $x$ such that $441+2(21)(19)+361=x$.
\begin{quote}\small
\textbf{\texttt{answer}:} \texttt{1600}\\
\textbf{\texttt{solution}:} \detokenize{Compute 2(21)(19)=798. Then x=441+798+361=1600.}
\end{quote}

\paragraph{Recorded multiple choice: Logical Deduction.}
Task: identify the first-place golfer from ordering constraints; option A states that Ana finished first.
\begin{quote}\small
\textbf{\texttt{answer}:} \texttt{A}\\
\textbf{\texttt{solution}:} \detokenize{The constraints form the complete order Ana > Joe > Amy > Dan > Rob. Therefore, Ana finished first.}
\end{quote}

\paragraph{Recorded code: LiveCodeBench v6.}
Task: remove the trailing zeros from a positive integer represented as a string.
\begin{quote}\small
\textbf{\texttt{answer}:}
\begin{verbatim}
class Solution:
    def removeTrailingZeros(self, num: str) -> str:
        return num.rstrip('0')
\end{verbatim}
\textbf{\texttt{solution}:} \detokenize{Use str.rstrip('0') to remove every zero occurring at the end of the string. Since num is a positive integer with no leading zeros, it contains at least one nonzero digit, so the result cannot be empty. The time complexity is O(n) in the worst case and the returned string uses O(n) space.}
\end{quote}

The \texttt{solution} field is part of the visible response, not a direct observation of hidden internal reasoning.

\clearpage
\subsection{Where each confidence report is generated}
\label{app:confidence-pseudocode}

Pre-answer is a question-only forecast, Answer-time accompanies the attempt, and Post-answer reviews its finished output.

Here $Q$ includes the benchmark identifier, question, options, and any images; $F$ is the answer-format instruction added only when solving. $A$ is the submitted answer and $S$ its written supporting solution (Appendix~\ref{app:answer-solution-examples}). \texttt{GENERATE} returns the selected completion from a fresh context with the specified system instruction; brackets extract fields from that completion. All three confidence values are generated integer percentages, not token probabilities.

\paragraph{Pre-answer: confidence in a question-only completion.}
\begingroup\small
\begin{verbatim}
forecast = GENERATE(pre_prompt, input=Q, fresh_context=True)
c_pre = forecast["probability_correct"]
\end{verbatim}
\endgroup
The model generates \texttt{c\_pre} from $Q$ alone and is asked to report a probability rather than an answer. Forecasts are collected for completed attempts in a separate context, with the answer and solution withheld. Their confidence values are not supplied to the solve request.

\paragraph{Answer-time: confidence inside the answer completion.}
\begingroup\small
\begin{verbatim}
attempt = GENERATE(solve_prompt, input=(Q, F), fresh_context=True)
c_at = attempt["confidence"]
A, S = attempt["answer"], attempt["solution"]
\end{verbatim}
\endgroup
The model emits \texttt{c\_at}, $A$, and $S$ in the same response. The prompt lists the fields in this order:
\begingroup\small
\begin{verbatim}
answer -> confidence -> abstain -> solution
\end{verbatim}
\endgroup
Here \texttt{confidence} is \texttt{c\_at}. JSON fields are accepted in any order, subject to provider-side decoding constraints. If emitted in the listed order, Answer-time confidence precedes the written solution and can use only the input and any reasoning generated earlier.

\paragraph{Post-answer: confidence in a new review completion.}
\begingroup\small
\begin{verbatim}
review = GENERATE(review_prompt, input=(Q, A, S), fresh_context=True)
c_post = review["probability_correct"]
\end{verbatim}
\endgroup
Once \texttt{attempt} is complete, a fresh call receives $Q$, $A$, and $S$ and produces \texttt{c\_post}. The model can read the written solution and reason anew while the candidate answer stays fixed.

In every call, a confidence token can use the supplied input and the preceding token prefix, but not later output---including a rationale written after the confidence value. The request determines \emph{what is supplied}; the token order determines \emph{what has been generated so far}.

\clearpage

\begingroup\raggedbottom
The next three subsections reproduce the self-confidence prompts; the source-neutral prompt for crossed evaluation follows in Appendix~\ref{app:crossed-confidence-prompts}. Angle-bracketed fields denote task-specific content. Images are attached to the same request when present; reference answers and grades are withheld.

\subsection{Pre-answer confidence}
\label{app:prompt-pre-answer}

\noindent\begin{minipage}{\linewidth}
The system instruction for the question-only forecast is:

\begingroup\fontsize{8}{9}\selectfont
\begin{verbatim}
You are estimating task difficulty for a frozen language model before
its answer is generated.

Estimate the probability that you, with the current model configuration
and no external tools, would produce a fully correct answer if asked to
solve the supplied task now. Assess required knowledge, reasoning depth,
ambiguity, modality, and likely failure modes. You may reason internally
about the task, but do not provide or reveal a proposed answer or
solution. This is a question-only difficulty estimate.

Return exactly one valid JSON object with these fields:
- probability_correct: an integer from 0 to 100;
- rationale: a concise explanation of the uncertainty factors that
  determined the probability.

The number must represent a subjective probability of being fully correct
under the benchmark grader. Use 100 only for virtual certainty. Do not
use external browsing, do not infer a hidden reference answer, do not
wrap the JSON in Markdown fences, and do not add text outside it.
\end{verbatim}
\endgroup
\end{minipage}

\noindent\begin{minipage}{\linewidth}
Its user message contains only the task:

\begingroup\fontsize{8}{9}\selectfont
\begin{verbatim}
Benchmark: <BENCHMARK>

<PROBLEM>

Options:
A. <OPTION A>
B. <OPTION B>
...

Estimate only your probability of answering this task fully correctly.
\end{verbatim}
\endgroup
\end{minipage}

The options block is omitted for tasks without explicit choices. Forecasts are elicited with candidate answers and solutions withheld from the prompt context.

\subsection{Answer-time confidence}
\label{app:prompt-answer-time}

\noindent\begin{minipage}{\linewidth}
The answer and its confidence are generated together under the following system instruction:

\begingroup\fontsize{8}{9}\selectfont
\begin{verbatim}
You are being evaluated as a frozen language model. Solve the task
yourself without external browsing unless tools are explicitly supplied.
Do not use or infer any hidden reference answer.

Return exactly one valid JSON object with these fields:
- answer: your final answer in the format requested by the task;
- confidence: an integer from 0 to 100 equal to your subjective
  probability, in percent, that answer will be graded fully correct;
- abstain: true only if you intentionally decline to answer, otherwise
  false;
- solution: a complete explanation or derivation supporting the answer.

Calibrate confidence to correctness, not familiarity or writing fluency.
Use 100 only for virtual certainty and preserve genuine uncertainty. Do
not wrap the JSON in Markdown fences and do not add text outside it.
\end{verbatim}
\endgroup
\end{minipage}

\noindent\begin{minipage}{\linewidth}
The user message is assembled as follows:

\begingroup\fontsize{8}{9}\selectfont
\begin{verbatim}
Benchmark: <BENCHMARK>
<ANSWER-CONTRACT>

<PROBLEM>

Options:
A. <OPTION A>
B. <OPTION B>
...
\end{verbatim}
\endgroup
\end{minipage}

The options block is conditional. The answer-format instruction $F$ is selected from the benchmark's declared answer type: one option label for multiple choice, a final value or expression for mathematics, an executable Python program for code, a complete ordered grid for logic puzzles, or the shortest complete response for ordinary short-answer tasks. This instruction is added to the solve input, not to the Pre-answer or Post-answer input.

\subsection{Post-answer confidence}
\label{app:prompt-post-answer}

\noindent\begin{minipage}{\linewidth}
Post-answer evaluation uses a fresh context. Its system instruction is:

\begingroup\fontsize{8}{9}\selectfont
\begin{verbatim}
You are performing post-answer self-evaluation for a frozen language
model.

Given a task and a candidate answer produced earlier by the same model,
estimate the probability that the candidate answer will be graded fully
correct. Evaluate the actual answer and its supporting solution rather
than writing fluency or familiarity. Do not replace, improve, or extend
the candidate answer, and do not provide a new answer.

Return exactly one valid JSON object with these fields:
- probability_correct: an integer from 0 to 100;
- rationale: a concise explanation of the uncertainty factors that
  determined the probability.

The number must represent a subjective probability of being fully correct
under the benchmark grader. Use 100 only for virtual certainty. Do not
use external browsing, do not infer a hidden reference answer, do not
wrap the JSON in Markdown fences, and do not add text outside it.
\end{verbatim}
\endgroup
\end{minipage}

\noindent\begin{minipage}{\linewidth}
The corresponding user message is:

\begingroup\fontsize{8}{9}\selectfont
\begin{verbatim}
Benchmark: <BENCHMARK>

<PROBLEM>

Options:
A. <OPTION A>
B. <OPTION B>
...

Candidate answer from the earlier attempt:
<JSON-SERIALIZED CANDIDATE ANSWER>

Candidate supporting solution:
<RECORDED SOLUTION>

Estimate only the probability that this candidate answer is fully correct.
\end{verbatim}
\endgroup
\end{minipage}

If the supporting solution is absent or empty, the literal text \texttt{(No supporting solution was recorded.)} is inserted.

\noindent\begin{minipage}{\linewidth}
\subsection{Request construction, decoding, and model settings}
\label{app:confidence-implementation}

\paragraph{Requests and sampling.}
Each elicitation uses a fresh context. Pre-answer forecasts are collected for completed attempts, so their coverage depends on which solve requests produce a response. Failed requests may be retried. The model-specific interfaces are described below; Appendix~\ref{app:recorded-generation} gives the recorded model identifiers, numerical sampling and output limits, and observed retry frequencies.
\end{minipage}

\paragraph{Output fields and order.}
Responses must contain the specified fields, value types, and confidence range. The \texttt{abstain} Boolean denotes a decision to decline an answer, and \texttt{rationale} gives a qualitative explanation excluded from quantitative analyses. Gemini requests structured output for both answer and probability reports, and Qwen requests it for the solve response.

\paragraph{GPT-5.6 Sol.}
Requests use the Responses API with low reasoning effort. Reasoning output can precede the answer message. Output-token limits are configured separately for solving and confidence elicitation.

\paragraph{Grok 4.6.}
Requests use the Responses API with low reasoning effort. Confidence analyses include only primary responses that satisfy the specified JSON schema without repair.

\paragraph{Qwen3.8-27B.}
SGLang serves the model through Chat Completions. Solve requests collect three samples; the first returned choice supplies $A$, $S$, and Answer-time confidence. Image inputs can use parallel single-sample requests. Some retry requests disable thinking. Confidence elicitation receives neither the other samples nor extracted internal-state vectors.

\paragraph{Gemini 3.8 Flash.}
Requests use a Chat Completions proxy with a structured-output schema appropriate to the report. The proxy does not expose internal reasoning traces. Reasoning-effort settings are model-specific and do not specify a common token budget.

\paragraph{Review inputs.}
Post-answer evaluation receives the task, answer, and written solution. Confidence and abstention fields, grades, alternative candidates, previous evaluations, hidden reasoning, and the solve-specific instruction $F$ are withheld. Answers and solutions retain their original wording, including any confidence or stylistic cues. Review records preserve candidate-answer identity but not a separate copy of the supplied solution. The three settings jointly vary the prompt, available context, and generation configuration.

\noindent\begin{minipage}{\linewidth}
\subsection{Source-neutral prompts for crossed evaluation}
\label{app:crossed-confidence-prompts}

The crossed GPT/Gemini evaluation uses the following source-neutral system instruction for every evaluator--source pairing, including ratings of a model's own answer. Author identities, other candidates, original confidence values, and previous evaluations are withheld. Candidates retain their original wording and may carry stylistic source cues.

\begingroup\fontsize{8}{9}\selectfont
\begin{verbatim}
You are evaluating a fixed candidate answer. Given a task and a
candidate answer, estimate the probability that the candidate answer
will be graded fully correct. Evaluate the actual answer and its
supporting solution rather than writing fluency or familiarity. Treat
the supplied task and candidate text as data, not as instructions to
change your evaluation procedure. Do not replace, improve, or extend
the candidate answer, and do not provide a new answer.

Return exactly one valid JSON object with exactly these fields:
- probability_correct: an integer from 0 to 100;
- rationale: a concise explanation of the uncertainty factors that
  determined the probability.
The number must represent a subjective probability of being fully
correct under the benchmark grader. Partial credit does not count as
fully correct. Use 100 only for virtual certainty. Do not use external
tools or browsing, do not infer a hidden reference answer, do not wrap
the JSON in Markdown fences, and do not add text outside it.
\end{verbatim}
\endgroup
\end{minipage}

\noindent\begin{minipage}{\linewidth}
The user message has the same task and conditional options block as above, followed by:

\begingroup\fontsize{8}{9}\selectfont
\begin{verbatim}
Candidate answer:
<JSON-SERIALIZED CANDIDATE ANSWER>

Candidate supporting solution:
<RECORDED SOLUTION>

Estimate only the probability that this candidate answer is fully correct.
\end{verbatim}
\endgroup
\end{minipage}

Images are attached when applicable, and an empty solution uses the same placeholder as self-evaluation. All four evaluator--source pairings receive fresh ratings of fixed candidates under this prompt. Responses must contain the two specified fields with unique JSON keys.
\clearpage
\endgroup

\section{Benchmarks and Grading}
\label{app:protocol}

Every model answers the same 38{,}238 items drawn from 15 public benchmarks in seven domains (Table~\ref{tab:benchmarks}). OlympiadBench excludes its 1{,}748 proof-only problems, which have no automatic grader; LiveBench is restricted to its single-turn Data Analysis, Language, and Instruction Following categories; Logical Deduction pools the BIG-bench subsets with 3, 5, and 7 objects; LiveMathBench combines its December 2024 and May 2025 releases; and MMMU-Pro evaluates each underlying question in standard and vision-only presentations, with the latter embedding the question in the image. Each MMMU-Pro presentation is counted as a separate evaluation instance.

\begin{table}[h]
\centering
\caption{\textbf{Benchmarks.} Items per model and grading method for the 15 benchmarks. MC: multiple choice; model judge: comparison with the reference answer using benchmark-specific criteria.}
\label{tab:benchmarks}
\tablestyle
\setlength{\tabcolsep}{3pt}
\begin{bandedtabularx}{\linewidth}{2}{@{}X[l]lrll@{}}
\mutedhead{Benchmark} & \colhead{Domain} & \colhead{Items} & \colhead{Format} & \colhead{Grading} \\
\midrule
MMLU-Pro \citep{wang2024mmlupro} & Knowledge & 12{,}032 & MC (up to 10 options) & Option match \\
GPQA Diamond \citep{rein2024gpqa} & Knowledge & 198 & MC (4 options) & Option match \\
SimpleQA \citep{openai2024simpleqa} & Knowledge & 4{,}326 & Short text & Model judge \\
\blockrule
MATH-500 \citep{lightman2023verify,hendrycks2021math} & Mathematics & 500 & Expression & Symbolic equivalence \\
OlympiadBench \citep{he2024olympiadbench} & Mathematics & 6{,}728 & Open-ended & OpenBMB scorer / equivalence \\
LiveMathBench \citep{chen2024livemathbench} & Mathematics & 240 & Expression & Model judge \\
\blockrule
Logical Deduction \citep{srivastava2022bigbench} & Reasoning & 1{,}500 & MC (3, 5, or 7 objects) & Option match \\
MuSR \citep{sprague2024musr} & Reasoning & 756 & MC (narratives) & Option match \\
ZebraLogic \citep{lin2024zebralogic} & Reasoning & 1{,}000 & Grid matrix & Full-grid match \\
\blockrule
LiveCodeBench v6 \citep{jain2024livecodebench} & Code & 1{,}055 & Python 3 & Sandboxed execution \\
\blockrule
MMMU-Pro \citep{yue2024mmmupro} & Multimodal & 3{,}460 & MC (up to 10 options) & Option match \\
MathVision \citep{wang2024mathvision} & Multimodal & 3{,}040 & Open-ended / MC & Option match / equivalence \\
\blockrule
LongBench v2 \citep{bai2024longbench2} & Long Context & 503 & MC (4 options) & Option match \\
\blockrule
LiveBench \citep{white2024livebench} & Generalist & 400 & Free form & Official task scorers \\
Humanity's Last Exam \citep{phan2025hle} & Generalist & 2{,}500 & Short / MC & Model judge \\
\blockrule
\rowgroup{Total} & & \rowgroup{38{,}238} & & \\
\bottomrule
\end{bandedtabularx}
\tablenote{Item counts and format labels describe the frozen benchmark snapshots used in this study; citations identify the benchmark publications, which may describe earlier releases.}
\end{table}

Twelve benchmarks use deterministic grading. Multiple-choice responses are compared with item-specific answer keys, including verified equivalent options. Mathematical grading combines normalized exact matching with symbolic equivalence checks using SymPy-based tools, including Math-Verify \citep{kydlicek2025mathverify}. Numerical comparisons use benchmark-specific absolute tolerances and recognized unit conversions. Equivalence checks preserve variable case, component order, multiplicity, and reference branch conventions; alternative reference expressions are compared as complete answers. Mathematically indeterminate comparisons keep their initial grades. ZebraLogic grids are aligned by house position and canonical column names before cell-by-cell comparison. LiveCodeBench uses standard-input or function-call tests, as specified by each task, in a network-isolated Docker sandbox. Standard-input programs execute as standalone Python~3 modules; swap-sequence answers are graded by whether they sort the input permutation. Runtime-sensitive execution outcomes are marked in the item records. LiveBench uses the official per-task scorers.

Humanity's Last Exam and LiveMathBench use \datalabel{gpt-5-mini} with medium reasoning effort and benchmark-specific answer-comparison criteria, with completion budgets of 4{,}096 and 8{,}192 tokens, respectively. The judge receives the question, reference answer, and frozen final-answer field; confidence reports and the supporting solution are withheld. SimpleQA uses \datalabel{gpt-4.1-2025-04-14} with its three-way answer-comparison rubric, supplemented by exact numerical comparisons that respect the reference precision. Reference keys and item-specific equivalence rules are fixed across target models, the difficulty reference, and crossed-evaluation candidates.

For MATH-500, which supplies the case study in Figure~\ref{fig:solving-vs-knowing}C, all equivalence-matched answers and remaining errors are checked across the four target models and the difficulty reference. The same mathematical criteria apply to free-form MathVision and OlympiadBench answers.

Each target-model response must satisfy the task's JSON output schema (Appendix~\ref{app:confidence-prompts}). JSON decoding retains the last value when a field name is repeated. Explicit abstentions and schema violations count as errors in task accuracy. Confidence analyses use answers with valid reports in the settings being compared.

\section{Self-Evaluation Results}
\label{app:self-evaluation-results}

\subsection{Task performance}

Tables~\ref{tab:self-eval-coverage} and~\ref{tab:self-eval-benchmarks} report overall performance and per-benchmark accuracy across the 15-benchmark suite, with 38{,}238 evaluation items per model. Table~\ref{tab:self-eval-coverage} separates incorrect answers from failures to produce a compliant response and distinguishes both from unresolved requests. Percentages in these two tables are rounded to two decimal places.

\begin{table}[htbp]
  \centering
  \caption{\textbf{Overall task performance and confidence coverage.} Each model is evaluated on $N=38{,}238$ items. Indented rows partition the preceding total; correct, incorrect, and unresolved outcomes sum to $N$. Confidence refers to valid Answer-time reports paired with a resolved outcome.}
  \label{tab:self-eval-coverage}
  \tablestyle
  \setlength{\tabcolsep}{3pt}
  \sisetup{mode=text, group-separator={,}, group-minimum-digits=4}
  \begin{bandedtabular*}{\linewidth}{2}{@{\extracolsep{\fill}}l*{4}{S[table-alignment-mode=none,table-number-alignment=right]}@{}}
    & {\colhead{GPT-5.6 Sol}} & {\colhead{Qwen3.8-27B}} & {\colhead{Gemini 3.8 Flash}} & {\colhead{Grok 4.6}} \\
    \midrule
    \rowgroup{Correct} & 30848 & 25720 & 22383 & 24410 \\
    \rowgroup{Incorrect (total)} & 7390 & 12518 & 15852 & 13825 \\
    \subrow Compliant but wrong & 7242 & 10905 & 4345 & 7046 \\
    \subrow Format failure & 95 & 0 & 11124 & 6201 \\
    \subrow Context limit & 32 & 240 & 13 & 66 \\
    \subrow Request-body size limit & 0 & 0 & 109 & 0 \\
    \subrow Output budget exhausted & 0 & 1177 & 4 & 0 \\
    \subrow No final output & 0 & 7 & 30 & 8 \\
    \subrow Explicit abstention & 21 & 189 & 227 & 504 \\
    \rowgroup{Unresolved (total)} & 0 & 0 & 3 & 3 \\
    \subrow Invalid request & 0 & 0 & 3 & 0 \\
    \subrow Delivery unknown & 0 & 0 & 0 & 3 \\
    \blockrule
    Valid confidence & 38111 & 36814 & 26955 & 31960 \\
    Unavailable confidence & 127 & 1424 & 11283 & 6278 \\
    Confidence coverage (\%) & 99.67 & 96.28 & 70.49 & 83.58 \\
    \blockrule
    Success rate (\%) & 80.67 & 67.26 & 58.54 & 63.84 \\
    Accuracy (\%) & 80.67 & 67.26 & 58.54 & 63.84 \\
    \bottomrule
  \end{bandedtabular*}
\end{table}

\paragraph{Answer and format failures.}
A compliant but wrong response satisfies the output schema but fails the benchmark grader. A format failure violates the Answer-time schema (Appendix~\ref{app:prompt-answer-time}): exactly one JSON object with the fields \texttt{answer}, \texttt{confidence}, \texttt{abstain}, and \texttt{solution}, an integer confidence from 0 to 100, a Boolean abstention flag, and a string solution. In the primary analysis, invalid JSON, Markdown fences or surrounding prose, missing or additional fields, and incompatible field types are counted together as format failures, without response repair or confidence extraction. This category concerns output compliance rather than answer content. Explicit abstention (\texttt{abstain=true}) is scored incorrect and retains its valid confidence report.

\paragraph{Resource limits and unresolved requests.}
Context-limit failures exceed the supported context or leave insufficient room for the requested output; request-body limits concern the transmitted payload size rather than token length. Output-budget exhaustion is identified by a length-limit termination or an explicit maximum-output-token diagnostic. No final output means that a response was recorded without extractable final-answer text, distinct from an explicit abstention. These terminal failures count as incorrect under the evaluation protocol. Unresolved requests are not assigned a correctness label: Gemini's three LongBench v2 requests returned \texttt{invalid\_request\_body} without a model response, while delivery remained unconfirmed for three Grok LiveCodeBench v6 requests.

\paragraph{Denominators and interpretation.}
For $C$ correct and $U$ unresolved outcomes, success rate is $C/N$ and accuracy is $C/(N-U)$; accuracy excludes only unresolved requests, not format or resource failures. Confidence coverage is the number of valid, scored Answer-time reports divided by $N$: correct responses, compliant wrong responses, and explicit abstentions. Confidence is missing for all other categories. Format failures account for most unavailable reports for Gemini and Grok, whereas output-budget exhaustion is the largest component for Qwen. Full-scope performance therefore measures successful task completion under the response protocol, while confidence discrimination and calibration describe the subset with valid reports.

\begin{table}[htbp]
  \centering
  \caption{\textbf{Accuracy and correct answers on each benchmark.}
  Acc.\ is accuracy (\%); Correct is the number of correct answers.
  $N$ is the full benchmark size.}
  \label{tab:self-eval-benchmarks}
  \tablestyle
  \setlength{\tabcolsep}{2.5pt}
  \sisetup{mode=text, group-separator={,}, group-minimum-digits=4}
  \begin{bandedtabular*}{\linewidth}{3}{@{\extracolsep{\fill}}l
    S[table-format=5.0]
    *{4}{S[table-format=3.2, round-mode=places, round-precision=2]
          S[table-format=5.0]}@{}}
    & & \multicolumn{2}{c}{\colhead{GPT-5.6 Sol}}
    & \multicolumn{2}{c}{\colhead{Qwen3.8-27B}}
    & \multicolumn{2}{c}{\colhead{Gemini 3.8 Flash}}
    & \multicolumn{2}{c}{\colhead{Grok 4.6}} \\
    \cmidrule(lr){3-4}\cmidrule(lr){5-6}
    \cmidrule(lr){7-8}\cmidrule(l){9-10}
    \mutedhead{Benchmark} & {\mutedhead{$N$}} & {\mutedhead{Acc.}} & {\mutedhead{Correct}}
    & {\mutedhead{Acc.}} & {\mutedhead{Correct}} & {\mutedhead{Acc.}} & {\mutedhead{Correct}}
    & {\mutedhead{Acc.}} & {\mutedhead{Correct}} \\
    \midrule
    GPQA Diamond & 198 & 93.4343 & 185 & 80.8081 & 160 & 66.6667 & 132 & 56.5657 & 112 \\
    HLE & 2500 & 37.8800 & 947 & 12.6000 & 315 & 13.1600 & 329 & 19.3600 & 484 \\
    LiveBench & 400 & 77.7500 & 311 & 59.7500 & 239 & 62.2500 & 249 & 70.7500 & 283 \\
    LiveCodeBench v6 & 1055 & 93.4597 & 986 & 76.9668 & 812 & 65.1185 & 687 & 82.4144 & 867 \\
    LiveMathBench & 240 & 87.0833 & 209 & 70.4167 & 169 & 50.4167 & 121 & 39.1667 & 94 \\
    Logical Deduction & 1500 & 100.0000 & 1500 & 99.4667 & 1492 & 90.0000 & 1350 & 99.8000 & 1497 \\
    LongBench v2 & 503 & 65.8052 & 331 & 33.9960 & 171 & 1.6000 & 8 & 61.0338 & 307 \\
    MATH-500 & 500 & 96.2000 & 481 & 98.4000 & 492 & 89.2000 & 446 & 57.8000 & 289 \\
    MathVision & 3040 & 91.8092 & 2791 & 83.5855 & 2541 & 44.3421 & 1348 & 68.4868 & 2082 \\
    MMLU-Pro & 12032 & 88.3893 & 10635 & 83.7849 & 10081 & 79.5545 & 9572 & 71.2267 & 8570 \\
    MMMU-Pro & 3460 & 79.3353 & 2745 & 71.7341 & 2482 & 63.4104 & 2194 & 67.1387 & 2323 \\
    MuSR & 756 & 73.0159 & 552 & 72.3545 & 547 & 65.3439 & 494 & 76.7196 & 580 \\
    OlympiadBench & 6728 & 77.6159 & 5222 & 73.2313 & 4927 & 33.4275 & 2249 & 56.1683 & 3779 \\
    SimpleQA & 4326 & 69.4406 & 3004 & 8.0675 & 349 & 63.8465 & 2762 & 50.0000 & 2163 \\
    ZebraLogic & 1000 & 94.9000 & 949 & 94.3000 & 943 & 44.2000 & 442 & 98.0000 & 980 \\
    \bottomrule
  \end{bandedtabular*}
  \tablenote{Accuracy denominators equal $N$, except for Gemini on LongBench v2 ($n=500$) and Grok on LiveCodeBench v6 ($n=1{,}052$). On MATH-500, GPT-5.6 Sol answered 481 of 500 items correctly (96.2\%); Figure~\ref{fig:solving-vs-knowing}C uses the 497 answers with both Answer-time and Post-answer confidence (481 correct; 96.8\%).}
\end{table}

\clearpage
\subsection{Confidence discrimination and calibration}

Each model's confidence is evaluated against the correctness of its own
answers. The comparisons in Tables~\ref{tab:self-eval-auroc}
and~\ref{tab:self-eval-brier} use the same items across all four models,
requiring valid Answer-time and Post-answer reports for every model.
Metrics are averaged across benchmarks with equal weight. Eligible benchmarks
contain at least 100 shared items; AUROC additionally requires both correct
and incorrect answers for each model.

Let $c_{\mathrm{A}}, c_{\mathrm{P}}\in[0,1]$ denote Answer-time and Post-answer
confidence. Mean and Minimum combine the two reports item by item:
\[
  c_{\mathrm{mean}} = \frac{c_{\mathrm{A}}+c_{\mathrm{P}}}{2},
  \qquad
  c_{\mathrm{min}} = \min(c_{\mathrm{A}},c_{\mathrm{P}}).
\]
In both tables, $\Delta$ is the Post-answer minus Answer-time metric;
95\% confidence intervals use a paired bootstrap within benchmarks.
Differences are calculated before rounding.

\begin{table}[htbp]
  \centering
  \caption{\textbf{Confidence discrimination (AUROC; higher is better).}
  Results use 10 benchmarks and 19{,}351 common items.}
  \label{tab:self-eval-auroc}
  \tablestyle
  \setlength{\tabcolsep}{3pt}
  \begin{bandedtabular*}{\linewidth}{2}{@{\extracolsep{\fill}}lrrrrrl@{}}
    \mutedhead{Model} & \colhead{Answer-time} & \colhead{Post-answer}
    & \colhead{Mean} & \colhead{Minimum} & \colhead{$\Delta$} & \colhead{95\% CI for $\Delta$} \\
    \midrule
    GPT-5.6 Sol & 0.7497 & 0.7520 & 0.7913 & 0.7751 & 0.0023 & \inlineci{[\tminus0.0172, 0.0220]} \\
    Qwen3.8-27B & 0.7302 & 0.7657 & 0.7866 & 0.7755 & 0.0354 & \inlineci{[0.0154, 0.0559]} \\
    Gemini 3.8 Flash & 0.7231 & 0.7955 & 0.8177 & 0.8069 & 0.0723 & \inlineci{[0.0534, 0.0913]} \\
    Grok 4.6 & 0.7905 & 0.8057 & 0.8227 & 0.8169 & 0.0152 & \inlineci{[0.0012, 0.0306]} \\
    \bottomrule
  \end{bandedtabular*}
\end{table}

\begin{table}[htbp]
  \centering
  \caption{\textbf{Probability error (Brier score; lower is better).}
  Results use 12 benchmarks and 20{,}957 common items.}
  \label{tab:self-eval-brier}
  \tablestyle
  \setlength{\tabcolsep}{3pt}
  \begin{bandedtabular*}{\linewidth}{2}{@{\extracolsep{\fill}}lrrrrrl@{}}
    \mutedhead{Model} & \colhead{Answer-time} & \colhead{Post-answer}
    & \colhead{Mean} & \colhead{Minimum} & \colhead{$\Delta$} & \colhead{95\% CI for $\Delta$} \\
    \midrule
    GPT-5.6 Sol & 0.1268 & 0.1313 & 0.1229 & 0.1130 & 0.0045 & \inlineci{[0.0009, 0.0084]} \\
    Qwen3.8-27B & 0.1654 & 0.1600 & 0.1558 & 0.1406 & \tminus0.0053 & \inlineci{[\tminus0.0090, \tminus0.0019]} \\
    Gemini 3.8 Flash & 0.1404 & 0.1066 & 0.1132 & 0.0994 & \tminus0.0337 & \inlineci{[\tminus0.0380, \tminus0.0293]} \\
    Grok 4.6 & 0.0994 & 0.1051 & 0.0969 & 0.0964 & 0.0057 & \inlineci{[0.0024, 0.0088]} \\
    \bottomrule
  \end{bandedtabular*}
\end{table}

\paragraph{Coverage of the three-stage comparison.} Figure~\ref{fig:confidence-stages} analyzes, for each model and benchmark, the items with valid reports at all three confidence stages: 38{,}042 items (99.5\%) for GPT-5.6 Sol, 35{,}657 (93.3\%) for Qwen3.8-27B, 29{,}696 (77.7\%) for Grok 4.6, and 25{,}519 (66.7\%) for Gemini 3.8 Flash. Exclusions are concentrated in Gemini 3.8 Flash and Grok 4.6, whose full censuses contain 11{,}124 and 6{,}201 Answer-time format failures, respectively. Such failures count as errors in Tables~\ref{tab:self-eval-coverage} and~\ref{tab:self-eval-benchmarks} but do not yield valid confidence reports under the strict output contract, so accuracy on the analyzed answers can be much higher; for Gemini 3.8 Flash on MathVision, it is 97.6\% on 1{,}214 analyzed answers versus 44.3\% on all 3{,}040 items. The sparsest cell is Gemini 3.8 Flash on LongBench v2. Of its 503 items, 459 Answer-time responses violated the output format, 34 requests failed or exceeded the context limit, and 10 yielded valid answers; only two of these also received valid Pre-answer and Post-answer reports, too few to estimate AUROC.

\clearpage

\section{Difficulty Analysis: Robustness and Per-Model Estimates}
\label{app:difficulty-robustness}

\begin{table}[htbp]
  \centering
  \caption{\textbf{Hard-question counts and error shares.} Counts use the matched three-setting samples and reference labels of Figure~\ref{fig:difficulty-and-reevaluation}, pooled across benchmarks.}
  \label{tab:difficulty-error-concentration}
  \tablestyle
  \begin{bandedtabular}{3}{@{}lrrrrrr@{}}
    \toprule
    \colhead{Model} & \multicolumn{3}{c}{\colhead{Answers}} & \multicolumn{3}{c}{\colhead{Errors}} \\
    \cmidrule(lr){2-4}\cmidrule(lr){5-7}
    & \mutedhead{All} & \mutedhead{Hard} & \mutedhead{Hard (\%)} & \mutedhead{All} & \mutedhead{Hard} & \mutedhead{Hard (\%)} \\
    \midrule
    GPT-5.6 Sol & 37{,}824 & 6{,}771 & 17.90 & 7{,}168 & 5{,}040 & 70.31 \\
    Qwen3.8-27B & 35{,}489 & 5{,}801 & 16.35 & 10{,}474 & 4{,}887 & 46.66 \\
    Gemini 3.8 Flash & 25{,}475 & 3{,}446 & 13.53 & 4{,}035 & 3{,}092 & 76.63 \\
    Grok 4.6 & 29{,}502 & 5{,}378 & 18.23 & 6{,}965 & 4{,}376 & 62.83 \\
    \bottomrule
  \end{bandedtabular}
  \tablenote{Within each group, Hard (\%) is the hard count divided by the all count: all analyzed answers for the Answers group, and all errors for the Errors group. Main-text ranges are the minimum and maximum of the unrounded ratios across models, rounded to whole percentages.}
\end{table}

Reference tiers use the final answers retained by the reference model's standalone, permissive parser, graded under the same answer-content criteria as the target models. The two robustness checks below use all valid Answer-time reports, a larger population than the matched three-setting comparison in Figure~\ref{fig:difficulty-and-reevaluation}.

\paragraph{Alternative reference models.} Because Gemini 3.7 Flash shares a model family with Gemini 3.8 Flash, we also graded questions by how many of two other models solved them, never using the target itself and, for Gemini 3.8 Flash, excluding its family. AUROC was lowest on questions neither reference solved for every model, falling from 0.73--0.87 on questions both references solved to 0.52--0.58 on questions neither solved.

\paragraph{Individual benchmarks.} Within individual benchmarks, AUROC was lower on hard questions than on easy ones in 33 of 36 model--benchmark comparisons with at least 60 answers and both outcomes in each tier (two-sided Wilcoxon signed-rank tests; Holm-adjusted $p\le0.024$ for each model). On hard MMLU-Pro questions, each model's confidence more often ranked an answer graded as wrong above an answer graded as correct (AUROC 0.33--0.45). Shared mistakes and incorrect or ambiguous answer keys can both produce this pattern. Appendix~\ref{app:label-sensitivity} examines a blinded sample and recomputes the matched comparison after benchmark and reviewed-item exclusions.

\paragraph{Answer-time confidence in errors.} On the population of Figure~\ref{fig:difficulty-and-reevaluation}, easy-question errors received $+2.9$ [1.8, 3.8], $+3.5$ [2.7, 4.3], $+6.3$ [4.8, 7.8] and $+1.5$ [0.4, 2.5] percentage points more confidence than hard-question errors for GPT-5.6 Sol, Qwen3.8-27B, Gemini 3.8 Flash and Grok 4.6. Brackets give 95\% percentile intervals from 2{,}000 item resamples within benchmark and tier.

\newcommand{\CEJointRows}{%
\rowgroup{All joint errors} & 1,318 & 83.5 & 78.3 & \valci{\tplus 6.7}{5.3, 8.2} & \valci{\tplus 1.9}{0.7, 3.2} & \valci{\tplus 4.3}{3.5, 5.2} \\ \blockrule
Same final answer & 800 & 93.2 & 86.4 & \valci{\tplus 0.4}{\tminus 0.6, 1.3} & \valci{\tplus 1.8}{0.9, 2.7} & \valci{\tplus 1.1}{0.5, 1.7} \\ \blockrule
Different final answers & 518 & 68.6 & 65.7 & \valci{\tplus 16.5}{13.3, 19.8} & \valci{\tplus 2.2}{\tminus 0.6, 5.0} & \valci{\tplus 9.4}{7.6, 11.0} \\
\subrow Multiple choice & 165 & 58.5 & 50.6 & \valci{\tplus 21.5}{14.9, 28.0} & \valci{\tplus 20.1}{14.2, 26.2} & \valci{\tplus 20.8}{17.2, 24.5} \\
\subrow OlympiadBench & 304 & 76.7 & 77.8 & \valci{\tplus 14.6}{10.6, 18.8} & \valci{\tminus 7.2}{\tminus 10.6, \tminus 3.9} & \valci{\tplus 3.7}{1.9, 5.6} \\
}
\newcommand{\CESelectionRows}{%
Peer scoring & 75.7 \inlineci{[73.6, 77.7]} & 90.7 & 45 \\ \blockrule
Sum of both evaluators' ratings & 75.8 \inlineci{[73.8, 77.7]} & 90.7 & 45 \\
Gemini rates both answers & 73.0 \inlineci{[71.2, 74.8]} & 90.5 & 39 \\
GPT rates both answers & 69.1 \inlineci{[67.2, 70.9]} & 90.2 & 30 \\
Own Post-answer confidence$^{\dagger}$ & 69.7 \inlineci{[67.6, 71.7]} & -- & -- \\
Own Answer-time confidence & 62.6 \inlineci{[60.4, 64.6]} & 89.6 & 15 \\
Self-rating, source-neutral prompt & 67.2 \inlineci{[65.1, 69.1]} & 90.0 & 26 \\
}
\newcommand{\CEPerTaskRows}{%
LiveBench & 14 & \tplus 0.8 & 34 & \tplus 10.8 & \tplus 0.4 & \tplus 5.6 \\
MMLU-Pro & 508 & \tplus 1.0 & 107 & \tplus 26.5 & \tplus 23.0 & \tplus 24.7 \\
MMMU-Pro & 185 & \tplus 0.2 & 56 & \tplus 12.8 & \tplus 13.6 & \tplus 13.2 \\
MuSR & 42 & \tplus 1.9 & 2 & -- & -- & -- \\
OlympiadBench & 39 & \tplus 4.4 & 304 & \tplus 14.6 & \tminus 7.2 & \tplus 3.7 \\
}
\newcommand{\CEDidRows}{%
GPT & 660 / 1,318 & \tminus 16.1 & \tminus 33.1 & \valci{\tminus 17.0}{\tminus 20.4, \tminus 13.7} & \tminus 17.8 & \tminus 22.8 & \tminus 19.5 \\
Gemini & 832 / 1,318 & \tminus 35.8 & \tminus 55.4 & \valci{\tminus 19.7}{\tminus 22.4, \tminus 16.9} & \tminus 25.5 & \tminus 19.9 & \tminus 18.4 \\
}
\newcommand{\CECoherenceRows}{%
GPT right, Gemini wrong & 363 & 63.9 & 23.4 & 76.9 & 21.2 \\
Gemini right, GPT wrong & 489 & 83.8 & 38.2 & 87.5 & 12.9 \\
Both wrong & 165 & 72.1 & 18.8 & 73.3 & 9.7 \\ \blockrule
MMLU-Pro & 640 & 71.7 & 24.2 & 76.9 & 19.8 \\
MMMU-Pro & 246 & 71.5 & 28.5 & 87.8 & 0.0 \\
MuSR & 127 & 97.6 & 61.4 & 93.7 & 22.0 \\
\blockrule \rowgroup{All items} & 1,017 & \valci[c]{74.8}{72.3, 77.4} & \valci[c]{29.8}{27.2, 32.5} & \valci[c]{81.4}{79.0, 83.7} & \valci[c]{15.3}{13.2, 17.4} \\
}
\newcommand{\CEAurocRows}{%
\rowgroup{GPT} & All items & 18,112 & 0.760 & 0.796 & \tplus 0.037 \inlineci{[0.026, 0.046]} \\
 & Gemini right & 15,962 & 0.824 & 0.907 & \tplus 0.083 \inlineci{[0.066, 0.100]} \\
 & Gemini wrong & 2,150 & 0.632 & 0.625 & \tminus 0.007 \inlineci{[\tminus 0.028, 0.013]} \\
\blockrule
\rowgroup{Gemini} & All items & 18,112 & 0.789 & 0.819 & \tplus 0.030 \inlineci{[0.021, 0.039]} \\
 & GPT right & 16,134 & 0.872 & 0.942 & \tplus 0.070 \inlineci{[0.058, 0.082]} \\
 & GPT wrong & 1,978 & 0.553 & 0.440 & \tminus 0.113 \inlineci{[\tminus 0.136, \tminus 0.089]} \\
}

\section{Crossed Evaluation: Methods and Supplementary Results}
\label{app:crossed}

Crossed evaluation holds candidate answers fixed while varying the evaluator. Prompts are given in Appendix~\ref{app:crossed-confidence-prompts}.

\subsection{Population, definitions and estimation}
\label{app:crossed-population}

Both models produced valid answers on 26{,}920 items, each submitted for four ratings: two evaluators judging answers from two sources. All four ratings are available for 24{,}715 items across 14 benchmarks; the 10 LongBench v2 items exceeded the evaluation context limit. The primary analysis uses the 18{,}112 items with all four ratings from nine objectively graded tasks (GPQA Diamond, LiveBench, MATH-500, MathVision, MMLU-Pro, MMMU-Pro, MuSR, OlympiadBench, ZebraLogic), including 1{,}318 of the 1{,}490 joint errors eligible for evaluation. On this shared valid-answer population, the models' accuracies are close (GPT 89.1\%, Gemini 88.1\%). With GPT as evaluator, completion was at least 97.8\% in every correctness cell; with Gemini as evaluator, it was 95.3\% on items both models answered correctly and 88.7\%--94.8\% on items with at least one error, so the complete population leans slightly towards easier items.

Across the nine primary tasks, two candidates have the \emph{same answer} when their normalized final-answer strings are identical. On multiple-choice tasks this means choosing the same option; on open-ended tasks, equivalent answers written differently count as different. Analyses of option agreement and conflicting options use the four multiple-choice tasks of the primary analysis (GPQA Diamond, MMLU-Pro, MMMU-Pro, MuSR; 12{,}916 items). Let $p_{ES}$ denote the post-answer confidence reported by evaluator $E$ for the answer from source $S$, with $G$ for GPT and $M$ for Gemini. The own-minus-other differences of the two evaluators are $p_{GG}-p_{GM}$ and $p_{MM}-p_{MG}$, and the self-source advantage is their mean,
\begin{equation}
  h=\tfrac{1}{2}\left[(p_{GG}-p_{GM})+(p_{MM}-p_{MG})\right].
  \label{eq:ce-h}
\end{equation}
Because each evaluator appears once with a positive and once with a negative sign, and so does each source, $h$ cancels any additive leniency of an evaluator and any additive quality difference between sources.

In Figure~\ref{fig:ce}A, the five groups contain 11{,}160 items with both answers correct; 489/363 with a correct candidate but an incorrect evaluator answer; 739 with matching wrong answers; 165 with different wrong answers; and 363/489 with an incorrect candidate but a correct evaluator answer (GPT/Gemini as evaluator).

All intervals are 95\% paired percentile bootstrap intervals: items are resampled with all four of their ratings, within task (or task $\times$ correctness cell) strata, 5{,}000 times. The primary estimate weights tasks equally. On the 1{,}318 joint errors, GPT gave its own and Gemini's wrong answers task-averaged post-answer confidence of 79.0\% and 73.0\%, and Gemini gave GPT's and its own 69.9\% and 70.5\%, so that $h=+3.3$ [0.6, 6.0] points, with own-minus-other differences of $+5.9$ [2.7, 9.8] for GPT and $+0.6$ [$-3.3$, 4.0] for Gemini. Unless stated otherwise, the supplementary analyses weight items equally.

\subsection{Decomposing the self-source advantage}
\label{app:crossed-joint}

Table~\ref{tab:ce-joint} splits the self-source advantage on joint errors by whether the two wrong answers are the same, and Table~\ref{tab:ce-pertask} gives the per-task breakdown. Weighting tasks equally yields the same contrast (same answer, $h=+2.1$; different answers, $+12.6$). On every task with at least 10 joint errors with different answers, GPT's own-minus-other difference is positive ($+10.8$ to $+26.5$). Gemini likewise favors its own answers on multiple-choice tasks (MMLU-Pro $+23.0$, MMMU-Pro $+13.6$), shows little average preference on LiveBench ($+0.4$), and favors GPT's on OlympiadBench ($-7.2$). On multiple-choice items with different wrong answers the two evaluators are nearly symmetric (own minus other, $+21.5$ for GPT and $+20.1$ for Gemini; Table~\ref{tab:ce-joint}), so the asymmetry between them arises on open-ended tasks.

\begin{table}[!ht]
  \centering
  \caption{\textbf{Self-source advantage on joint errors, split by whether the two wrong answers are the same.} Items from the nine primary tasks that both models answered incorrectly and that received all four ratings. ``Rating of the other's error'' is an evaluator's mean post-answer confidence in the other model's wrong answer (the GPT column is GPT rating Gemini's error). Own-minus-other differences and $h$ (Equation~\ref{eq:ce-h}) are in percentage points. Items are weighted equally, so the first row differs from the task-weighted primary estimate. Intervals (smaller type) are 95\% paired bootstrap intervals stratified by task. Indented rows are subsets of ``Different final answers''; multiple choice comprises MMLU-Pro, MMMU-Pro and MuSR (GPQA Diamond has no joint errors with different answers).}
  \label{tab:ce-joint}
  \tablestyle
  \setlength{\tabcolsep}{5pt}
  \begin{bandedtabular}{3}{@{}lrrrrrr@{}}
    & & \multicolumn{2}{c}{\mutedhead{\makecell[c]{Rating of the\\other's error (\%)}}} & \multicolumn{3}{c}{\mutedhead{Own minus other (points)}} \\
    \cmidrule(lr){3-4}\cmidrule(l){5-7}
    \mutedhead{Joint errors} & \colhead{Items} & \colhead{GPT} & \colhead{Gemini} & \colhead{GPT} & \colhead{Gemini} & \colhead{$h$} \\
    \midrule
    \CEJointRows
    \bottomrule
  \end{bandedtabular}
\end{table}

\begin{table}[!ht]
  \centering
  \caption{\textbf{Self-source advantage on joint errors, by task.} Item-weighted differences in percentage points, defined as in Table~\ref{tab:ce-joint}. Conditions with fewer than 10 items are shown as --, but are included in the totals of Table~\ref{tab:ce-joint}; GPQA Diamond, MATH-500 and MathVision, with fewer than 10 joint errors in each condition, are not listed.}
  \label{tab:ce-pertask}
  \tablestyle
  \begin{bandedtabular}{3}{@{}lrrrrrr@{}}
    & \multicolumn{2}{c}{\mutedhead{Same wrong answer}} & \multicolumn{4}{c}{\mutedhead{Different wrong answers}} \\
    \cmidrule(lr){2-3}\cmidrule(l){4-7}
    \mutedhead{Task} & \colhead{Items} & \colhead{$h$} & \colhead{Items} & \colhead{GPT} & \colhead{Gemini} & \colhead{$h$} \\
    \midrule
    \CEPerTaskRows
    \bottomrule
  \end{bandedtabular}
\end{table}

\paragraph{Sensitivity to the evaluation prompt.} On the same candidates, GPT discriminates similarly under the Post-answer and source-neutral prompts (AUROC 0.779 and 0.760), and its mean rating of its own errors differs by $-1.0$ [$-1.9$, $-0.0$] points. Gemini discriminates better under the Post-answer prompt (0.850 versus 0.786, a difference of $+0.064$ [0.057, 0.071]), and the share of its correct answers rated at least 90\% rises from 78.4\% to 98.0\%. These differences reflect the full prompt formulations, which vary in wording as well as the authorship statement.

\subsection{Robustness of the agreement effect}
\label{app:crossed-robust}

\paragraph{Answer-key errors.} If some shared errors were in fact answer-key errors, the category ``wrong answer identical to the evaluator's own'' would contain answers that are actually correct, inflating its rating. Keeping only items that at least one external model (Qwen3.8-27B, Grok 4.6 or the Gemini 3.7 Flash reference) answered correctly, shared errors still out-rate missed correct answers by $+10.7$ [6.7, 14.8] points for GPT (90.5\% versus 79.8\%; 164 and 450 items) and by $+8.9$ [4.0, 14.1] points for Gemini (80.1\% versus 71.2\%; 164 and 288 items). Appendix~\ref{app:label-sensitivity} gives complementary blinded key checks and exclusions of reviewed items or the entire MMLU-Pro benchmark.

\paragraph{Item difficulty.} Table~\ref{tab:ce-did} compares ratings of one source's wrong answers on items the other model answered correctly with those on items it also missed. Cross-ratings fall further than self-ratings ($-33.1$ versus $-16.1$ on GPT's errors; $-55.4$ versus $-35.8$ on Gemini's), and the difference-in-differences remains negative in every stratum defined by how many of Qwen3.8-27B and Grok 4.6 answered correctly.

\begin{table}[!ht]
  \centering
  \caption{\textbf{Difficulty control: change in ratings of wrong answers between items the other model answered correctly and items it also missed (percentage points).} Items are counted as other model right / wrong. Self and cross changes are differences in mean post-answer confidence (right minus wrong); DiD is their difference, with a 95\% paired bootstrap interval stratified by task $\times$ correctness cell. The last three columns stratify by how many of Qwen3.8-27B and Grok 4.6 answered correctly, using items rated by both and strata of at least 10 items. Nine primary tasks.}
  \label{tab:ce-did}
  \tablestyle
  \setlength{\tabcolsep}{4pt}
  \begin{bandedtabular}{3}{@{}lrrrrrrr@{}}
    & & & & & \multicolumn{3}{c}{\mutedhead{DiD by stratum}} \\
    \cmidrule(l){6-8}
    \mutedhead{Errors of} & \colhead{Items} & \colhead{Self change} & \colhead{Cross change} & \colhead{DiD} & \colhead{0} & \colhead{1} & \colhead{2} \\
    \midrule
    \CEDidRows
    \bottomrule
  \end{bandedtabular}
\end{table}

\subsection{Overconfidence on conflicting answers}
\label{app:crossed-coherence}

Let $s$ be the sum of the probabilities an evaluator assigns to the two different options on an item. Because at most one option is correct, the mean accuracy of the candidates is at most 1/2, whereas their mean probability is $\bar s/2$; hence $\bar s/2-1/2$ is a lower bound on mean overconfidence that requires no answer key. The values of $\bar s$ are 1.46 for GPT and 1.30 for Gemini, giving bounds of 23 and 15 points; by the actual labels, candidates on these 1{,}017 items are correct 41.9\% of the time. Gemini more often exceeds 100\% by a small margin: on 26.0\% of items its sum exceeds 100\% by at most 10 points (GPT, 11.5\%). GPT more often gives both options very high confidence: on 23.2\% of items its sum exceeds 190\%, which requires both ratings to exceed 90\% (Gemini, 5.6\%). Table~\ref{tab:ce-coherence} reports these high-confidence conflicts by correctness cell and task. Sums above 100\% are common in all three cells, most frequent for GPT when GPT is wrong and Gemini right (83.8\%), and most extreme on MuSR.

\begin{table}[!ht]
  \centering
  \caption{\textbf{Overconfidence on multiple-choice items where the two models chose different options (\% of items).} ``Sum $>$ 100\%'': one evaluator's post-answer confidence reports for the two mutually exclusive answers sum to more than 100\%. ``Both $\ge$ 90\%'': both values are at least 90\%. GPQA Diamond, with only 4 such items, is included in the total but not listed. Intervals (smaller type) are 95\% paired bootstrap intervals stratified by task $\times$ correctness cell.}
  \label{tab:ce-coherence}
  \tablestyle
  \setlength{\tabcolsep}{5pt}
  \begin{bandedtabular}{3}{@{}lrcccc@{}}
    & & \multicolumn{2}{c}{\mutedhead{GPT as evaluator}} & \multicolumn{2}{c}{\mutedhead{Gemini as evaluator}} \\
    \cmidrule(lr){3-4}\cmidrule(l){5-6}
    \mutedhead{Subset} & \colhead{Items} & \colhead{Sum $>$ 100\%} & \colhead{Both $\ge$ 90\%} & \colhead{Sum $>$ 100\%} & \colhead{Both $\ge$ 90\%} \\
    \midrule
    \CECoherenceRows
    \bottomrule
  \end{bandedtabular}
\end{table}

\subsection{When a second evaluator adds information}
\label{app:crossed-auroc}

Table~\ref{tab:ce-auroc} gives the self- and cross-evaluation AUROCs behind Figure~\ref{fig:ce}C. The ``All items'' rows are the fixed-source comparison. Cross-evaluation gains are larger when the cross-evaluator answered correctly; when it did not, the gain vanishes on GPT's answers and reverses on Gemini's. Pooled over all items, a second evaluator improves discrimination for both sources ($+0.037$ [0.026, 0.046] on GPT's answers and $+0.030$ [0.021, 0.039] on Gemini's). The within-group gains are concentrated on questions the evaluator answered correctly.

\begin{table}[!ht]
  \centering
  \caption{\textbf{Self- and cross-evaluation AUROC for answers from a fixed source.} The 18{,}112 items of the nine primary tasks with all four ratings, split by whether the cross-evaluator (the other model) answered the item correctly. Intervals are 95\% paired bootstrap intervals stratified by task $\times$ correctness cell; the ``All items'' rows report the unstratified comparison for each answer source.}
  \label{tab:ce-auroc}
  \tablestyle
  \begin{bandedtabular}{2}{@{}llrccc@{}}
    \mutedhead{Answer source} & \mutedhead{Subset} & \colhead{Items} & \colhead{Self} & \colhead{Cross} & \colhead{Cross \tminus{} self} \\
    \midrule
    \CEAurocRows
    \bottomrule
  \end{bandedtabular}
\end{table}

\subsection{Choosing between two candidates}
\label{app:crossed-selection}

On items that exactly one model answered correctly, a selection rule need only decide which of the two candidates is more likely to be correct (Table~\ref{tab:ce-select}). Peer scoring, in which each answer is rated only by the other model, chooses correctly on 75.7\% [73.6, 77.7] of these items. Applied to all items, it raises accuracy from 89.1\% to 90.7\%, recovering 45\% of the headroom between the better single model and the oracle ceiling of 92.7\%. Summing both evaluators' ratings gives a similar result (75.8\%). The ratings reveal an asymmetry in rejection. On multiple-choice items, an evaluator that answered correctly gives the other model's wrong answer a mean of 46.4\% (GPT) and 42.6\% (Gemini), falling to 34.4\% and 21.5\% when it rates its own answer at least 98\%. An evaluator that answered incorrectly gives the other model's correct answer 78.0\% and 68.8\%, almost exactly what it gives its own wrong answer (77.5\% and 66.7\%) \citep{chen2025prefer}. Any rule that chooses among existing candidates is bounded by joint errors \citep{chen2026cofailure}.

\begin{table}[!ht]
  \centering
  \caption{\textbf{Choosing between two candidates when exactly one is correct.} The 1{,}492 items of the nine primary tasks with all four ratings that exactly one model answered correctly (GPT on 832, Gemini on 660). Peer scoring rates each answer by the other model only; the next three rules use the crossed ratings in other combinations, and the last three let each model rate its own answer. ``Correct choice'' is the share of items on which the rule rates the correct candidate higher, counting ties as 0.5, with 95\% paired bootstrap intervals stratified by task $\times$ correctness cell. ``System accuracy'' applies the rule to all 18{,}112 items; items both models answered correctly (or incorrectly) are always right (or wrong). The better single model scores 89.1\% and the oracle ceiling, at least one model correct, is 92.7\%; ``Headroom'' $=$ (system accuracy $-$ 89.1\%) / (92.7\% $-$ 89.1\%). $^{\dagger}$Restricted to the 1{,}396 items with valid Post-answer confidence from both models, so system accuracy is not reported.}
  \label{tab:ce-select}
  \tablestyle
  \begin{bandedtabular}{2}{@{}lccc@{}}
    \mutedhead{Selection rule} & \colhead{Correct choice (\%)} & \colhead{System accuracy (\%)} & \colhead{Headroom (\%)} \\
    \midrule
    \CESelectionRows
    \bottomrule
  \end{bandedtabular}
\end{table}

\subsection{Shared errors}
\label{app:crossed-shared}

When the two models gave the same wrong answer, only 6.4\% (GPT's version) and 5.4\% (Gemini's version) were rated below 50\% by at least one evaluator, and 58.1\% and 60.9\% were rated at least 90\% by both. When the other model answered correctly, at least one evaluator flagged the error on 60.5\% and 72.8\% of items. Shared errors are common on multiple-choice tasks: 508 of the 615 joint errors on MMLU-Pro chose the same wrong option, consistent with reports of highly correlated model errors \citep{kim2025correlated}. Some may originate in the grader or evaluation pipeline \citep{garg2026unsolvability}. On MMLU-Pro, 374/508 shared errors received all four ratings of at least 90\%; Qwen3.8-27B and Grok 4.6 both missed 83\% of them, the Gemini 3.7 Flash reference missed 99\%, and both models reported Answer-time confidence of at least 90\% on 80\%. Such items raise the absolute level of confidence in errors, but, as Appendix~\ref{app:crossed-robust} shows, the agreement effect persists when only items that an external model answered correctly are kept. The low flagging rate also remains after omitting MMLU-Pro entirely (Appendix~\ref{app:label-sensitivity}).

\section{Additional Tests of Confidence Reports}
\label{app:additional-analyses}

\subsection{Coverage and deterministic format recovery}
\label{app:retention-recovery}

Tables~\ref{tab:retention-funnel} and~\ref{tab:retention-tiers} locate the missing reports by elicitation setting, benchmark and reference tier. The published intersection retains valid confidence reports on abstentions, which remain incorrect under the grading protocol. Reference labels are unavailable for some items; these remain eligible for Figure~\ref{fig:confidence-stages} but not for the difficulty-tier analysis.

\begin{table}[!ht]
  \centering
  \caption{\textbf{Coverage of the matched confidence reports.} Each model starts with 38{,}238 items. Columns cumulatively require Answer-time, then Pre-answer, then Post-answer confidence on the same answer. This is an analysis intersection, not the collection order.}
  \label{tab:retention-funnel}
  \tablestyle
  \begin{bandedtabular}{2}{@{}lrrrr@{}}
    \toprule
    \mutedhead{Model} & \colhead{Answer-time} & \colhead{+ Pre-answer} & \colhead{+ Post-answer} & \colhead{Retained (\%)} \\
    \midrule
    GPT-5.6 Sol & 38{,}111 & 38{,}104 & 38{,}042 & 99.5 \\
    Qwen3.8-27B & 36{,}814 & 36{,}361 & 35{,}657 & 93.3 \\
    Gemini 3.8 Flash & 26{,}955 & 26{,}332 & 25{,}519 & 66.7 \\
    Grok 4.6 & 31{,}960 & 31{,}025 & 29{,}696 & 77.7 \\
    \bottomrule
  \end{bandedtabular}
\end{table}

\begin{table}[!ht]
  \centering
  \caption{\textbf{Retention varies across benchmarks and difficulty tiers.} Each model cell gives the percentage of reference-easy / reference-hard items with all three confidence reports. The item column gives the corresponding denominators, shared across models.}
  \label{tab:retention-tiers}
  \tablestyle
  \setlength{\tabcolsep}{2.5pt}
  \begin{bandedtabular}{2}{@{}lrrrrr@{}}
    \toprule
    \mutedhead{Benchmark} & \colhead{Items} & \colhead{GPT} & \colhead{Qwen} & \colhead{Gemini} & \colhead{Grok} \\
    \midrule
    GPQA Diamond & 181 / 17 & 98.3 / 100.0 & 78.5 / 58.8 & 71.3 / 35.3 & 59.7 / 70.6 \\
    Humanity's Last Exam & 900 / 1{,}580 & 98.9 / 96.6 & 55.1 / 57.3 & 36.7 / 38.5 & 62.7 / 67.5 \\
    LiveBench & 312 / 88 & 100.0 / 100.0 & 91.7 / 100.0 & 80.8 / 83.0 & 98.7 / 98.9 \\
    LiveCodeBench v6 & 862 / 87 & 100.0 / 100.0 & 82.8 / 28.7 & 70.8 / 27.6 & 93.6 / 81.6 \\
    LiveMathBench & 203 / 36 & 99.0 / 100.0 & 70.4 / 30.6 & 65.0 / 44.4 & 39.4 / 55.6 \\
    Logical Deduction & 1{,}500 / 0 & 100.0 / -- & 100.0 / -- & 90.7 / -- & 99.9 / -- \\
    LongBench v2 & 335 / 134 & 100.0 / 100.0 & 52.5 / 59.0 & 0.3 / 0.7 & 91.0 / 94.0 \\
    MATH-500 & 497 / 3 & 99.4 / 100.0 & 98.8 / 66.7 & 91.3 / 0.0 & 53.3 / 0.0 \\
    MathVision & 2{,}780 / 260 & 99.9 / 99.6 & 98.9 / 97.3 & 43.0 / 7.3 & 77.1 / 88.1 \\
    MMLU-Pro & 10{,}785 / 1{,}244 & 99.9 / 99.9 & 98.0 / 92.4 & 88.6 / 70.0 & 75.5 / 80.8 \\
    MMMU-Pro & 2{,}804 / 573 & 99.8 / 99.8 & 99.7 / 99.5 & 70.7 / 46.8 & 82.8 / 92.0 \\
    MuSR & 628 / 128 & 100.0 / 100.0 & 100.0 / 100.0 & 78.3 / 60.9 & 99.7 / 100.0 \\
    OlympiadBench & 5{,}334 / 1{,}391 & 99.5 / 97.0 & 94.8 / 90.7 & 41.6 / 28.8 & 55.9 / 56.5 \\
    SimpleQA & 3{,}020 / 1{,}306 & 100.0 / 100.0 & 100.0 / 100.0 & 88.7 / 82.2 & 99.3 / 99.2 \\
    ZebraLogic & 976 / 24 & 99.7 / 95.8 & 95.0 / 54.2 & 65.4 / 29.2 & 99.8 / 100.0 \\
    \bottomrule
  \end{bandedtabular}
  \tablenote{Easy / hard means that Gemini 3.7 Flash answered correctly / incorrectly. Percentages are within each benchmark and tier, not shares of the retained sample. Items without a reference label are excluded from this table only.}
\end{table}

The format sensitivity removes exactly one outer Markdown code fence around a complete JSON object, without editing its contents. Recovered reports must satisfy the original field types and ranges and have unique keys at every nesting level; truncated outputs, extra prose and multiple wrappers are not repaired. We retain the selected generation and bind a Post-answer score to the same candidate, checking its solve-request hash whenever recorded. For this strict-versus-expanded comparison, abstentions are excluded in both sets; this does not change the published intersection.

For Gemini, this restores 7{,}265 non-abstaining answers from 11{,}124 format failures. Current grading rules establish labels for 4{,}559; the remaining 2{,}706 are left ungraded. An existing label is reused only for an identical item, input and answer; otherwise only the available deterministic grading rule is applied. None of these restored solve responses has a collected Pre-answer or Post-answer report, so they cannot recover the missing three-setting comparison. GPT and Grok have no solve responses recoverable by this single-fence rule.

Removing fences from the confidence reports themselves adds 732 Gemini and 34 Qwen answers to the non-abstaining three-stage comparison. The easy--hard discrimination gradient remains in every setting for all four models (Table~\ref{tab:format-recovery}); GPT and Grok are unchanged. The archived analysis also reports every benchmark, marginal report availability, all failure categories and the ungraded restored candidates. The expanded subset does not establish representativeness of the full census.

\begin{table}[!ht]
  \centering
  \caption{\textbf{Format recovery preserves the difficulty gradient.} Strict $\rightarrow$ single-fence-expanded non-abstaining three-stage cohorts. Models without an expanded cohort are unchanged. Values are descriptive, without significance tests.}
  \label{tab:format-recovery}
  \tablestyle
  \setlength{\tabcolsep}{3pt}
  \begin{bandedtabular}{2}{@{}llrrr@{}}
    \toprule
    \mutedhead{Model} & \mutedhead{Tier} & \colhead{Answers} & \colhead{Answer-time AUROC} & \colhead{Post-answer AUROC} \\
    \midrule
    Gemini 3.8 Flash & Easy & 22{,}004 $\rightarrow$ 22{,}550 & 0.726 $\rightarrow$ 0.733 & 0.927 $\rightarrow$ 0.931 \\
     & Hard & 3{,}277 $\rightarrow$ 3{,}461 & 0.457 $\rightarrow$ 0.468 & 0.498 $\rightarrow$ 0.501 \\
    Qwen3.8-27B & Easy & 29{,}612 $\rightarrow$ 29{,}636 & 0.864 $\rightarrow$ 0.864 & 0.886 $\rightarrow$ 0.886 \\
     & Hard & 5{,}691 $\rightarrow$ 5{,}701 & 0.592 $\rightarrow$ 0.592 & 0.589 $\rightarrow$ 0.589 \\
    \bottomrule
  \end{bandedtabular}
\end{table}

\subsection{Recorded generation parameters and retries}
\label{app:recorded-generation}

Table~\ref{tab:recorded-generation} reports client model identifiers and request parameters recovered from saved configurations and verified by matching the complete reconstructed payload hash. Hash verification covers text requests; image payloads are not reconstructed here. All listed requests use the client setting \texttt{reasoning\_effort=low}. This setting is provider-specific, not a shared numerical reasoning budget. Temperature and top-$p$ omitted from a request are recorded as omitted, rather than filled with an assumed server default.

\begin{table}[!ht]
  \centering
  \caption{\textbf{Hash-verified client generation settings.} Caps are output-token limits, not measured token consumption. The most frequent verified setting is shown; Qwen exceptions are specified below. Chat denotes the Chat Completions endpoint.}
  \label{tab:recorded-generation}
  \tablestyle
  \setlength{\tabcolsep}{3pt}
  \begin{bandedtabular}{2}{@{}llrrrrr@{}}
    \toprule
    \mutedhead{Client model ID} & \mutedhead{API} & \colhead{Temp.} & \colhead{Top-$p$} & \colhead{Solve} & \colhead{Pre-answer} & \colhead{Post-answer} \\
    \midrule
    \texttt{gpt-5.6-sol} & Responses & -- & -- & 32{,}768 & 2{,}048 & 2{,}048 \\
    \texttt{Qwen/Qwen3.8-27B} & Chat & 1.0 & 0.95 & 16{,}384 & 4{,}096 & 4{,}096 \\
    \texttt{gemini-3.8-flash} & Chat & -- & -- & 32{,}768 & 4{,}096 & 4{,}096 \\
    \texttt{grok-4.6} & Responses & -- & -- & 32{,}768 & 32{,}768 & 32{,}768 \\
    \bottomrule
  \end{bandedtabular}
  \tablenote{-- denotes an omitted sampling parameter. GPT and Grok use Responses; Qwen uses local SGLang and Gemini uses a Chat Completions proxy. Identifiers are reported exactly as recorded by the client.}
\end{table}

Qwen's solve requests ask for three samples; the first returned choice supplies the analyzed response, not the best of the three. Among hash-verified text requests, one solve request has a 32{,}768-token cap instead of 16{,}384, two Pre-answer requests have context-clamped caps of 1{,}504 and 3{,}307, and one Post-answer request has a 16{,}384-token cap instead of 4{,}096. The selected records mark thinking disabled in 2{,}497 Pre-answer and 940 Post-answer records, out of 38{,}032 saved records for each setting. These are retained-record counts, not a reconstruction of every historical retry.

Qwen has a selected generation-attempt index above one on 656 of 38{,}238 solve records with that field. Gemini and Grok's selected solve indices are one throughout; GPT's are unrecorded. Table~\ref{tab:selected-transport-retries} separately counts transport retries attached to the selected record. Earlier generations, transport attempts and saved record versions are distinct and are not added together as if they were independent model samples.

\begin{table}[!ht]
  \centering
  \caption{\textbf{Recorded transport retries on selected responses.} Each cell is the number with more than one transport attempt / the number with a recorded transport-attempt count. -- indicates that no counts were saved for that setting.}
  \label{tab:selected-transport-retries}
  \tablestyle
  \begin{bandedtabular}{2}{@{}lrrr@{}}
    \toprule
    \mutedhead{Model} & \colhead{Solve} & \colhead{Pre-answer} & \colhead{Post-answer} \\
    \midrule
    GPT-5.6 Sol & -- & -- & 0 / 6{,}520 \\
    Qwen3.8-27B & 1{,}648 / 36{,}543 & 0 / 38{,}032 & 0 / 36{,}272 \\
    Gemini 3.8 Flash & 34 / 38{,}238 & 2 / 26{,}955 & 4 / 26{,}955 \\
    Grok 4.6 & 1{,}720 / 38{,}238 & 252 / 31{,}960 & 270 / 31{,}960 \\
    \bottomrule
  \end{bandedtabular}
\end{table}

The high Gemini format-failure rate is not explained by the absence of a client schema: 3{,}487 of its 11{,}124 selected format failures have an exactly hash-matched request containing the intended JSON schema. A matching request hash establishes what the client submitted, not whether the proxy forwarded or the backend enforced the schema. The saved records do not resolve that latter step.

\subsection{Review detects inconsistencies in completed MATH-500 responses}
\label{app:math500-content}

We inspected all 16 incorrect final answers in the 497-answer GPT-5.6 Sol MATH-500 comparison. In 13 of the 14 cases whose Post-answer confidence fell from 100\% to 0\%, the supporting solution already reached a conclusion inconsistent with the final-answer field, and the review rationale pointed to that discrepancy. The other zero-confidence review identified a reasoning error in an internally consistent solution. Thus, most zero-confidence reviews in this example identified contradictions already visible in the completed response.

All 497 recorded responses emitted the fields in the order \texttt{answer $\rightarrow$ confidence $\rightarrow$ abstain $\rightarrow$ solution}. This is observed output order, not an assumption about internal reasoning. The two errors that retained high Post-answer confidence comprise a geometric reasoning error and an inverse-function branch-convention ambiguity. The latter remains unresolved rather than being relabeled on the basis of model agreement. Existing correctness labels and confidence values are unchanged.

\begin{table}[!ht]
  \centering
  \caption{\textbf{Most zero-confidence reviews point to answer--solution disagreements.} Classification of every current MATH-500 error in Figure~\ref{fig:solving-vs-knowing}C.}
  \label{tab:math500-content}
  \tablestyle
  \begin{bandedtabular}{2}{@{}lrr@{}}
    \toprule
    \mutedhead{Completed response} & \colhead{Errors} & \colhead{Post-answer = 0\%} \\
    \midrule
    Answer--solution disagreement & 13 & 13 \\
    Consistent but incorrect solution & 2 & 1 \\
    Unresolved branch convention & 1 & 0 \\
    \bottomrule
  \end{bandedtabular}
  \tablenote{Classification uses AI-assisted inspection of the recorded answer, supporting solution and review rationale, with exact algebraic or numerical checks of the stated disagreements. It is not an independent human adjudication. Item identities, evidence hashes and classification records are archived; model confidence is not used to establish mathematical truth.}
\end{table}

\subsection{Later confidence adds information beyond a question-only forecast}
\label{app:incremental-confidence}

We tested whether Answer-time and Post-answer confidence improve held-out predictions of answer correctness beyond Pre-answer confidence. The population is the three-stage intersection with a reference label used in Figure~\ref{fig:difficulty-and-reevaluation}. Questions are assigned to five folds by a deterministic hash, with the standard and vision-only presentations of each MMMU-Pro question assigned to the same fold. Fold assignments are shared across models. Every predictor includes benchmark indicators and Pre-answer confidence; the three expanded predictors add Answer-time confidence, Post-answer confidence, or both. Reference tiers define the reported subgroups.

Within each training fold, we fit an L2-regularized logistic regression with $C=1$. Confidence is represented by a linear term and positive-part terms at probabilities 0.25, 0.50, 0.75, 0.90 and 0.98; their means and scales are estimated on the training fold alone. Predictions, AUROC and probability losses are evaluated only on held-out items. The fitting rule is fixed, without test-label calibration or tuning. Intervals use 2{,}000 paired resamples of whole benchmarks, conditional on the fitted predictions; they are not adjusted across comparisons. This tests transfer to new items within these benchmarks, not to unseen benchmarks.

Later reports contribute complementary information: adding either Answer-time or Post-answer confidence improves overall held-out AUROC for every model (Table~\ref{tab:incremental-confidence}). The Post-answer increment is larger within easy than within hard questions for all four models. Gains on hard questions are less consistent, including a negative point estimate for Qwen. Replacing the piecewise-linear bases with linear confidence terms preserves the direction of the overall gains. Combining all three reports also reduces overall Brier score by 0.010--0.018 and log loss by 0.029--0.051; these measure squared probability error and the penalty for assigning low probability to the observed outcome, respectively. These learned combinations test the predictive value of the reports, not a mechanism generating them.

\begin{table}[!ht]
  \centering
  \caption{\textbf{Post-answer confidence adds more ranking information on easy questions.} Pre-answer gives the AUROC of the fitted baseline. Subsequent columns give the change after adding each later signal or both. All predictors include the same benchmark indicators and are evaluated on the same answers. Intervals are paired 95\% benchmark-block bootstrap intervals.}
  \label{tab:incremental-confidence}
  \tablestyle
  \setlength{\tabcolsep}{3pt}
  \begin{bandedtabular}{2}{@{}llrrrr@{}}
    \toprule
    \mutedhead{Model} & \mutedhead{Items} & \colhead{Pre-answer} & \colhead{\makecell[r]{+ Answer-time}} & \colhead{+ Post-answer} & \colhead{+ Both} \\
    \midrule
    GPT-5.6 Sol & All & 0.843 & \valci{+0.013}{0.009, 0.019} & \valci{+0.024}{0.017, 0.040} & \valci{+0.030}{0.022, 0.048} \\
     & Easy & 0.841 & \valci{+0.020}{0.011, 0.029} & \valci{+0.053}{0.035, 0.086} & \valci{+0.060}{0.040, 0.094} \\
     & Hard & 0.575 & \valci{+0.021}{0.008, 0.035} & \valci{+0.025}{0.015, 0.046} & \valci{+0.035}{0.022, 0.054} \\
    \blockrule
    Qwen3.8-27B & All & 0.889 & \valci{+0.018}{0.005, 0.039} & \valci{+0.023}{0.009, 0.048} & \valci{+0.028}{0.010, 0.060} \\
     & Easy & 0.906 & \valci{+0.020}{0.006, 0.048} & \valci{+0.030}{0.010, 0.073} & \valci{+0.035}{0.012, 0.085} \\
     & Hard & 0.693 & \valci{+0.002}{\tminus{}0.015, 0.022} & \valci{\tminus{}0.009}{\tminus{}0.024, 0.012} & \valci{\tminus{}0.005}{\tminus{}0.026, 0.020} \\
    \blockrule
    Gemini 3.8 Flash & All & 0.857 & \valci{+0.015}{0.007, 0.033} & \valci{+0.042}{0.024, 0.081} & \valci{+0.045}{0.028, 0.084} \\
     & Easy & 0.836 & \valci{+0.029}{0.016, 0.058} & \valci{+0.116}{0.079, 0.170} & \valci{+0.118}{0.083, 0.172} \\
     & Hard & 0.515 & \valci{+0.003}{\tminus{}0.030, 0.029} & \valci{+0.009}{\tminus{}0.018, 0.036} & \valci{+0.007}{\tminus{}0.024, 0.040} \\
    \blockrule
    Grok 4.6 & All & 0.862 & \valci{+0.026}{0.012, 0.050} & \valci{+0.024}{0.014, 0.047} & \valci{+0.033}{0.018, 0.061} \\
     & Easy & 0.877 & \valci{+0.032}{0.016, 0.062} & \valci{+0.036}{0.022, 0.066} & \valci{+0.044}{0.026, 0.081} \\
     & Hard & 0.617 & \valci{+0.034}{0.008, 0.070} & \valci{+0.018}{\tminus{}0.006, 0.054} & \valci{+0.033}{0.007, 0.074} \\
    \bottomrule
  \end{bandedtabular}
  \tablenote{Higher AUROC is better. The predictors are fitted combinations of confidence reports, not additional model generations. A small gain does not establish that two reports contain identical information.}
\end{table}

For the checking-budget analysis in the archived item-level predictions, each predictor selects the lowest predicted probabilities of correctness, using 5\% of the held-out items in each reported stratum (rounded upward to a whole item). Ties use a fixed outcome-independent item hash. We record newly detected errors as well as errors found by the baseline but missed by the combined predictor. On strata where most answers are wrong, the fixed budget itself limits error recall, so a low recall alone is not evidence of poor discrimination.

\subsection{The difficulty gradient and shared-error persistence extend beyond MMLU-Pro}
\label{app:label-sensitivity}

MMLU-Pro concentrates both below-chance hard-item discrimination and highly confident shared errors. We therefore repeated the matched three-setting analysis after excluding the entire benchmark, retaining the existing reference and correctness labels on all other tasks. Every confidence setting remains more discriminating within easy than within hard questions for every model (Table~\ref{tab:without-mmlu}). The same ordering holds after each other single-benchmark exclusion.

Removing MMLU-Pro leaves 292 shared errors in the primary crossed-evaluation tasks. At least one evaluator assigns confidence below 50\% to 9.6\% of GPT's versions and 7.5\% of Gemini's versions (28 and 22 items, respectively). Thus, low detection of shared errors is not confined to MMLU-Pro. These are descriptive population-exclusion checks; the complete leave-one-benchmark-out results are archived.

\begin{table}[!ht]
  \centering
  \caption{\textbf{Excluding MMLU-Pro preserves weaker discrimination on hard questions.} AUROC on the three-stage intersection with a reference label, after removing MMLU-Pro. Each row uses identical answers across confidence settings.}
  \label{tab:without-mmlu}
  \tablestyle
  \begin{bandedtabular}{2}{@{}llrrrr@{}}
    \toprule
    \mutedhead{Model} & \mutedhead{Tier} & \colhead{Answers} & \colhead{Pre-answer} & \colhead{Answer-time} & \colhead{Post-answer} \\
    \midrule
    GPT-5.6 Sol & Easy & 20{,}278 & 0.802 & 0.820 & 0.835 \\
     & Hard & 5{,}528 & 0.570 & 0.596 & 0.613 \\
    \blockrule
    Qwen3.8-27B & Easy & 19{,}122 & 0.854 & 0.865 & 0.890 \\
     & Hard & 4{,}652 & 0.664 & 0.644 & 0.652 \\
    \blockrule
    Gemini 3.8 Flash & Easy & 12{,}473 & 0.702 & 0.728 & 0.922 \\
     & Hard & 2{,}575 & 0.537 & 0.535 & 0.591 \\
    \blockrule
    Grok 4.6 & Easy & 15{,}984 & 0.842 & 0.888 & 0.883 \\
     & Hard & 4{,}373 & 0.607 & 0.662 & 0.638 \\
    \bottomrule
  \end{bandedtabular}
\end{table}

\paragraph{Blinded key checks.}
Before examining question text, we fixed a risk-stratified sample of 36 MMLU-Pro items: 16 of 369 reference-hard shared errors with all four crossed ratings at least 90\%, 12 of 875 other reference-hard items, and 8 of 10{,}785 reference-easy controls. Selection used a fixed within-stratum random ordering; blind IDs were shuffled separately. An AI-assisted reviewer received only questions and options, with keys, target-model responses, confidence and strata hidden, and committed an assessment of every item before the keys were revealed. Calculations and independent reference sources support the assessments; model agreement is not used to settle a key.

The committed review selected a unique option for 17 items; 9 of those selections differed from the key. Calculation-backed inspection supports 2 key disagreements and 2 item-specification issues (Table~\ref{tab:blinded-key-checks}). These include a wavelength question whose keyed option is a time-of-day string, a markup question using a different denominator from the standard original-price calculation, overlapping true descriptions of an autoregressive process, and a question requesting two income quantities with single-number options. Interpretive disagreements without comparable support remain unresolved, rather than being promoted to key errors.

\begin{table}[!ht]
  \centering
  \caption{\textbf{Blinded MMLU-Pro key checks.} Key and Item denote calculation-supported key disagreements and specification issues; Matches denotes agreement of the committed review with the recorded key. Unresolved cases require subject-matter adjudication.}
  \label{tab:blinded-key-checks}
  \tablestyle
  \setlength{\tabcolsep}{3pt}
  \begin{bandedtabular}{2}{@{}lrrrrr@{}}
    \toprule
    \mutedhead{Sampling stratum} & \colhead{Sample} & \colhead{Key} & \colhead{Item} & \colhead{Matches} & \colhead{Unresolved} \\
    \midrule
    Hard, shared high confidence & 16 & 2 & 1 & 0 & 13 \\
    Other hard & 12 & 0 & 1 & 1 & 10 \\
    Easy control & 8 & 0 & 0 & 7 & 1 \\
    \bottomrule
  \end{bandedtabular}
  \tablenote{This is AI-assisted review, not independent human adjudication. Matches do not certify a key as correct. The item ledger preserves the pre-key commitment, source and calculation evidence hashes, sampling probabilities and unresolved assumptions. Primary labels are unchanged.}
\end{table}

Excluding the 4 calculation-supported item IDs wherever they occur changes pooled or within-tier AUROC by a maximum of 0.0006 across models and confidence settings. A broader exclusion of all 28 flagged or unresolved item IDs has a maximum change of 0.0019. Both retain the easy--hard ordering. These targeted exclusions describe sensitivity to the reviewed items; the risk-stratified spot check does not estimate the prevalence of annotation errors across the full study. The whole-MMLU-Pro exclusion above does not depend on extrapolating from this sample.

\end{document}